\documentclass{article}

\PassOptionsToPackage{numbers,compress}{natbib}

\usepackage[main,final]{neurips_2026}

\usepackage[utf8]{inputenc} 
\usepackage[T1]{fontenc}    
\usepackage{hyperref}       
\usepackage{url}            
\usepackage{booktabs}       
\usepackage{amsfonts}       
\usepackage{nicefrac}       
\usepackage{microtype}      
\usepackage{amssymb}
\usepackage{amsmath}
\usepackage{multirow}
\usepackage{colortbl}
\usepackage[table]{xcolor}
\usepackage{graphicx}
\usepackage{tikz}
\usepackage{titlesec}
\usepackage{wrapfig}
\usepackage{algorithm}
\usepackage{algpseudocode}

\titlespacing*{\section}{0pt}{4pt plus 1pt minus 1pt}{3pt plus 1pt minus 1pt}
\titlespacing*{\subsection}{0pt}{2pt plus 1pt minus 1pt}{1pt plus 0.5pt minus 0.5pt}

\title{ViCoR: Estimating Visual Necessity via Counterfactual Residuals for Multimodal Medical Data Selection}

\author{%
Haotian Gan$^{1}$ \quad
Yudong Li$^{1,*}$ \quad
Xintian Li$^{1}$ \quad
Jiarui Han$^{2}$ \quad
Weidong Tang$^{3}$ \\[6pt]
$^{1}$Tsinghua University \hspace{1.3em}
$^{2}$The University of Hong Kong \hspace{1.3em}
$^{3}$China Agricultural University \\[3pt]
}

\begin{document}
\maketitle

\begingroup
\renewcommand{\thefootnote}{\fnsymbol{footnote}}
\footnotetext[1]{Corresponding author: \texttt{liyudong@tsinghua.edu.cn}}
\endgroup

\begingroup
\renewcommand{\thefootnote}{}
\footnotetext{Haotian Gan conducted this work during an internship at Tsinghua University.}
\endgroup

\begin{abstract}
Medical vision-language models often exhibit hidden stratification, where high overall performance masks failures on subtle visual findings and overlapping clinical conditions. We trace this failure mode to modality imbalance during fine-tuning, where strong linguistic priors can suppress visually grounded learning and induce plausible but weakly grounded rationales. Existing data selection methods usually score image-text pairs as holistic units, without separating visually necessary samples from shortcut-solvable ones. To address this problem, we propose ViCoR, a counterfactual-residual-based data selection method for Med-VLM fine-tuning. ViCoR estimates visual necessity through image-dependent optimization residuals, calibrates the resulting scores with medical supervision compatibility, and constructs a diversity-aware fine-tuning subset. Across two backbones, four external generalization benchmarks, and one multi-disease robustness benchmark, ViCoR-selected 20\% subsets achieve the strongest average performance compared with full-data fine-tuning and selection baselines. Gradient-level analysis and perturbation audits further show that ViCoR enriches image-dependent training signals and strengthens image-level and evidence-region dependence. The code will be available at https://github.com/htgan-ai/ViCoR.
\end{abstract}

\section{Introduction}
Despite the state-of-the-art performance of Medical Vision-Language Models (Med-VLMs) on existing benchmarks~\cite{wu2025medreason, liu-etal-2024-medcot, chen-etal-2024-m3cot}, their reliability in clinical settings remains a challenge~\cite{hartsock2024vision}.
One important issue is hidden stratification~\cite{oakden2020hidden}, where models achieve high overall accuracy while performing poorly on clinically important subgroups. 
This risk is especially consequential in high-stakes clinical scenarios where subtle visual findings determine the diagnosis.
For Med-VLMs, such subgroup failures are not limited to rare diseases. They can also arise from \textbf{underrepresented combinations of common diseases}~\cite{jeong2023optimized, seo2023enhancing} and from \textbf{ambiguous or non-standard visual presentations}~\cite{brown2023detecting, jin2024hidden}.
In these cases, visual evidence is often harder to interpret, which increases the risk of shortcut learning~\cite{geirhos2020shortcut}.
Instead of grounding predictions in visual evidence, models may rely on textual priors or disease co-occurrence patterns learned from biased training data~\cite{dancette2021beyond, niu2021counterfactual}. 
Recent work also shows that VLMs can over-trust textual priors and ignore uncertain visual evidence~\cite{deng2025words}. These findings raise broader concerns about apparent visual understanding and readiness in medical VLMs~\cite{asadi2026mirage,gu2025illusion}.

Such shortcut-driven predictions can create false trust when models generate plausible explanations not grounded in visual evidence~\cite{turpin2023language,moll2025evaluating}. 
One underlying cause is gradient starvation, where signals that are easier to learn suppress weaker but clinically important signals~\cite{pezeshki2021gradient}. 
In multimodal training, text often provides stronger optimization signals than images, which can slow or even stall the learning of the visual encoder~\cite{kwon-etal-2025-see}. 
This imbalance becomes more severe when training data contains many templated medical narratives. As a result, the model may learn to follow common textual patterns rather than use visual findings.

Since this imbalance is partly shaped by the fine-tuning distribution, data selection offers a practical way to adjust the learning signal.
However, existing data selection methods such as LESS~\cite{xia2024less}, Adapt-$\infty$~\cite{maharana2025adaptinfty}, and Self-Filter~\cite{chen2024your} estimate sample utility from influence matching, multimodal selection, or instruction quality filtering.
They usually treat each example as a whole image-text pair and do not distinguish whether its utility comes from visual evidence or from easy textual cues. 
This limitation is especially important in medical VLM fine-tuning, where text-dominated utility scores can favor templated diagnostic narratives and suppress visual evidence needed for diagnosis.
Recent work on Visual Information Gain (VIG) similarly selects samples according to how visual degradation changes prediction loss~\cite{lee2026focusing}. This motivates visual-aware data selection.
A more targeted selection criterion should therefore measure whether the image itself changes the learning signal, rather than only whether the sample appears difficult, influential, or well formatted~\cite{Rakib_2025_ICCV}.

\begin{figure*}[t]
    \centering
    \includegraphics[width=0.98\textwidth]{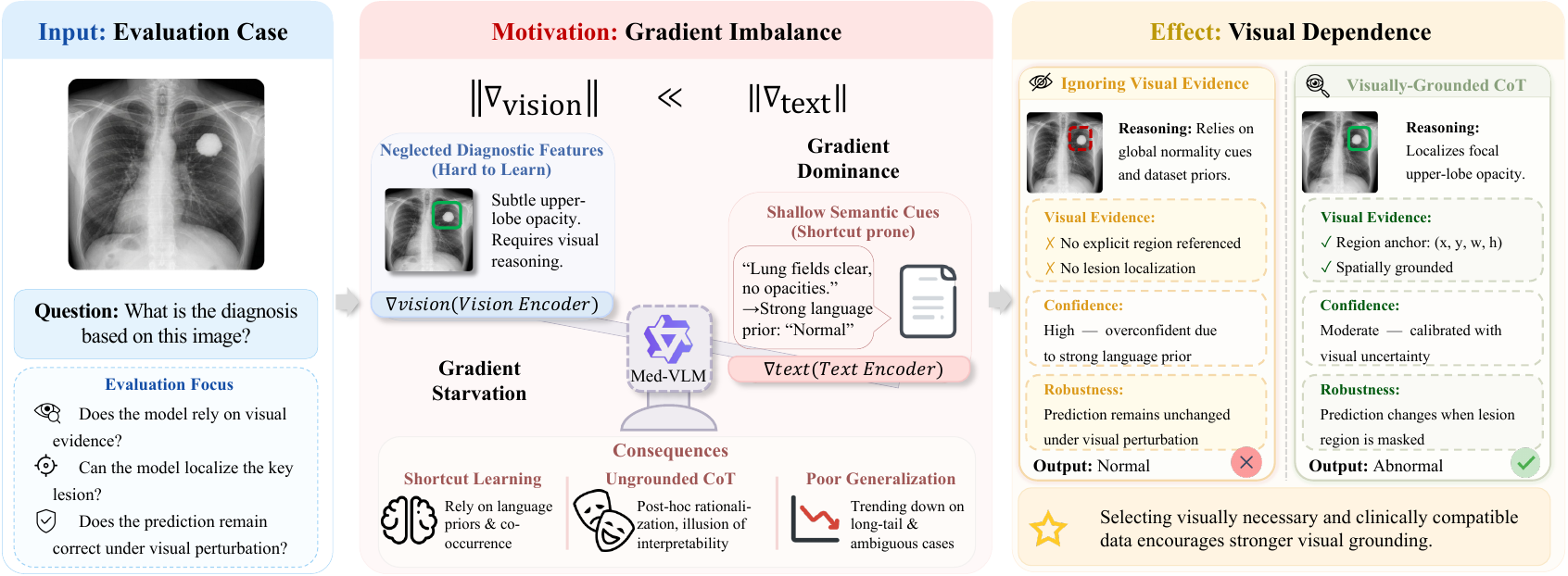}
    \vspace{-0.2em}
    \caption{Motivation and effect of ViCoR. ViCoR selects samples whose optimization signals depend on visual evidence, encouraging more visually grounded medical reasoning.}
    \label{fig:overview}
\end{figure*}

To address this limitation, we propose \textbf{ViCoR, a counterfactual-residual-based data selection method for Med-VLM fine-tuning.} As illustrated in Figure~\ref{fig:overview}, ViCoR mitigates gradient imbalance by selecting samples whose optimization signals depend on sample-specific visual evidence. It estimates visual necessity by comparing two gradients: one from the original image-question pair, and the other from a shortcut environment where the image is replaced by a mean-image shortcut control. Their difference estimates the image-dependent update not explained by textual shortcuts alone. Based on this score, ViCoR further calibrates supervision compatibility and builds a diversity-aware subset for fine-tuning. Our contributions are as follows:

\begin{itemize}
    \item We propose an image-dependent optimization residual that estimates visual necessity by isolating the gradient component induced by sample-specific visual evidence.
    \item We develop ViCoR, an offline selector that integrates residual-based scoring, medical supervision calibration, and diversity-aware subset construction.
    \item Across two backbones, four external generalization benchmarks, and one multi-disease robustness benchmark, ViCoR-selected 20\% subsets perform best on average. Perturbation audits further support stronger image and evidence-region dependence.
\end{itemize}

\section{Related Work}
\subsection{Faithfulness and Hallucination Evaluation}

Despite rapid progress in Med-VLMs, recent evaluations show that high benchmark scores do not ensure reliable reasoning. 
Models may reach correct answers through pattern matching or shortcut correlations rather than grounded visual reasoning~\cite{wang2025illusion, zhou2026drvdbench}.
This lack of grounded reasoning is also reflected in visual grounding, where models often fail in anatomical localization and relative spatial understanding~\cite{nguyen2025localizing, wolf2025your}.
Similar problems also appear in Chain-of-Thought reasoning and hallucination-sensitive clinical tasks, where models can generate plausible explanations or descriptions that are not well supported by the image~\cite{jiang2026m3cotbench, khanal2025hallucination}. 
These findings suggest that current Med-VLMs still tend to rely on linguistic priors, especially when visual evidence is ambiguous or difficult to use.

\subsection{Structured Reasoning and Clinical Alignment}
To mitigate hallucinations, prior work has mainly introduced structured inference, preference alignment, and causal intervention. For structured inference, Le et al.~\cite{le2025s} propose S-Chain to enforce visual reasoning steps, while Kim et al.~\cite{kim2025medclm} use curriculum learning for progressively complex diagnosis. Zang et al.~\cite{zang2025medical} further introduce logical decision trees. For preference alignment, Cui et al.~\cite{cui2024biomedical} use clinician-guided instruction tuning, while Liang et al.~\cite{liang2025chexpo} align outputs with clinical standards via counterfactual preference optimization. Yang et al.~\cite{yang2025walking} use counterfactual trajectories to control concept drift, and Zhao et al.~\cite{zhao2026aligning} improve consistency between findings and diagnoses. For causal intervention, Patil et al.~\cite{patil2023debiasing} weaken textual priors through causal minimization, while Li et al.~\cite{li-etal-2025-treble} use counterfactual samples for cross-modal learning. Ye et al.~\cite{ye2024causal} further target modality preference bias. However, Amit et al.~\cite{amit2025quantifying} show that modality imbalance remains persistent, with textual priors continuing to dominate visual signals in the training distribution.

\subsection{Data-Centric Curation and Selection}
Data-centric methods improve model behavior by changing the training data. Existing methods estimate data utility from different signals. Static difficulty methods such as IFD~\cite{li2024quantity} prune data based on instruction complexity. Optimization-based methods such as GraNd~\cite{paul2021deep} use gradient norms to measure sample importance, while ICONS~\cite{wu2024icons} aggregates influence scores to estimate utility more stably. ELFS~\cite{zheng2025elfs} builds label-free coresets to improve feature-space diversity, and XMAS~\cite{naharas2025data} aligns training trajectories to handle modality dynamics. However, these methods do not distinguish visual evidence from textual shortcuts when estimating data value. This distinction is crucial in medical VLM fine-tuning, where subtle lesions and localized findings often provide diagnostic cues beyond textual context. In contrast, ViCoR uses counterfactual residuals to favor samples whose optimization signals depend on visual evidence. Unlike holistic utility-based selection, ViCoR asks whether removing sample-specific visual content changes the trainable update, making the criterion explicitly modality-aware.

\section{Methodology}
\label{sec:methodology}

\begin{figure*}[t]
    \centering
    \includegraphics[width=0.98\textwidth]{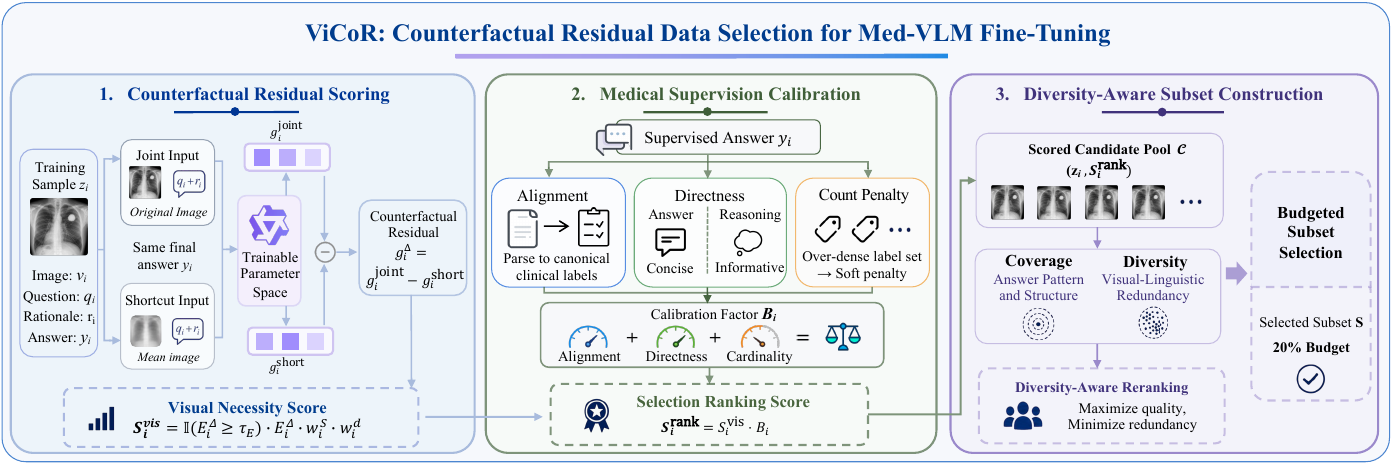}
    \vspace{-0.2em}
    \caption{Overview of ViCoR. The pipeline estimates counterfactual residual gradients, calibrates supervision quality, and applies diversity-aware construction to produce the final fine-tuning subset.}
    \label{fig:approach}
\end{figure*}

We propose ViCoR, a counterfactual-residual-based data selection method for Med-VLM fine-tuning. Figure~\ref{fig:approach} summarizes its three stages: estimating image-dependent residuals, calibrating them with medical supervision compatibility, and constructing a diversity-aware fine-tuning subset.

\subsection{Selection Objective and Scoring Setup}
\label{sec:problem}

Given a training pool $\mathcal{D}=\{z_i\}_{i=1}^{N}$ and a target budget $K$, ViCoR aims to select a subset $\mathcal{S}\subset\mathcal{D}$ with $|\mathcal{S}|=K$. Each example is denoted as $z_i=(v_i,q_i,r_i,y_i)$, with image $v_i$, question $q_i$, optional rationale prefix $r_i$, and final answer $y_i$. For datasets without rationale supervision, $r_i$ is set to an empty prefix.

We operationalize visual necessity as an optimization-level criterion: replacing the original image with a mean-image shortcut control should change the gradient update if the image contributes beyond the question and available rationale prefix. Since medical responses may include long rationales, ViCoR computes the scoring loss only on final-answer tokens:
\begin{equation}
\ell_i(\tilde v)
=
-\frac{1}{L_i}
\sum_{t=1}^{L_i}
\log p_\theta(y_{i,t}\mid \tilde v, q_i, r_i, y_{i,<t})
\end{equation}
where $y_i=(y_{i,1},\ldots,y_{i,L_i})$ denotes the final-answer token sequence and $\tilde v$ denotes the visual input used during scoring. The rationale prefix is kept identical in the joint and shortcut environments, so the residual measures how visual input changes the final-answer update under the same rationale-supervised context. If the rationale already makes the final answer predictable without image-specific evidence, the joint and shortcut gradients tend to be similar and yield a small residual.

All scores are computed offline before fine-tuning. For each candidate sample, gradients are computed only over the parameters that will be updated in the subsequent fine-tuning run, denoted as $\theta_s$. This makes the residual score reflect the sample's expected contribution to the actual fine-tuning update.

\subsection{Counterfactual Visual Residual}
\label{sec:visual_residual}

To estimate visual necessity, ViCoR compares two gradients in the same trainable parameter space. The \emph{joint environment} uses the original image and question, while the \emph{shortcut environment} keeps the question and supervised target unchanged but replaces the image with a neutral visual control. This design preserves textual priors, answer regularities, and the multimodal input format while removing sample-specific visual content:
\begin{equation}
g_i^{\mathrm{joint}}
=
\nabla_{\theta_s}\ell_i(v_i),
\qquad
g_i^{\mathrm{short}}
=
\nabla_{\theta_s}\ell_i(\bar v)
\end{equation}
where $\bar v$ denotes a neutral visual control implemented as a dataset-level mean image. The shortcut gradient estimates the update without sample-specific visual evidence, and the residual below captures the update that changes when the original image is restored. The image-dependent residual update is defined as
\begin{equation}
g_i^{\Delta}
=
g_i^{\mathrm{joint}}-g_i^{\mathrm{short}}
\end{equation}
If an example can be fitted primarily through language shortcuts, the joint and shortcut gradients should be similar in the same trainable parameter space, yielding a small residual. If the image changes the optimization signal, the residual becomes larger and indicates stronger visual necessity.

We measure the residual and shortcut strengths using gradient-norm scores computed from the token-normalized scoring loss:
\begin{equation}
E_i^{\Delta}
=
\|g_i^{\Delta}\|_2,
\qquad
E_i^{s}
=
\|g_i^{\mathrm{short}}\|_2
\end{equation}
$E_i^{\Delta}$ measures image-dependent residual strength, and $E_i^{s}$ measures the shortcut update under the same token-normalized scoring objective.

Residual strength alone does not indicate whether the visual signal competes with strong shortcuts. We therefore define a bounded shortcut dominance weight:
\begin{equation}
r_i
=
\frac{E_i^{s}+\epsilon}{E_i^{\Delta}+\epsilon},
\qquad
w_i^{s}
=
\log\left(1+\Pi_{[\rho_{\min},\rho_{\max}]}(r_i)\right)
\end{equation}
where $\Pi_{[a,b]}(x)=\min(\max(x,a),b)$ clips $x$ to $[a,b]$ to avoid unstable ratios. Larger $w_i^{s}$ indicates stronger shortcut dominance relative to the residual.

We also measure whether the residual update points in a direction that differs from the shortcut update:
\begin{equation}
w_i^{d}
=
\left[
1-\operatorname{cos}_{\epsilon}
\left(g_i^{\Delta},g_i^{\mathrm{short}}\right)
\right]_+
\end{equation}
where $[x]_+=\max(x,0)$ and
\begin{equation}
\operatorname{cos}_{\epsilon}(u,v)
=
\frac{u^\top v}{\|u\|_2\|v\|_2+\epsilon_{\cos}}
\end{equation}
This term emphasizes residuals that deviate from, rather than reinforce, shortcut-induced gradients.

The visual necessity score of ViCoR is then
\begin{equation}
S_i^{\mathrm{vis}}
=
\mathbb{I}\!\left(E_i^{\Delta}\ge \tau_E\right)
\cdot
E_i^{\Delta}
\cdot
w_i^{s}
\cdot
w_i^{d}
\end{equation}
Here, $\tau_E$ is a residual-strength threshold. The indicator term sets the score to zero when the image-dependent residual is negligible. For the remaining samples, $E_i^{\Delta}$ captures residual strength, $w_i^s$ captures shortcut pressure, and $w_i^d$ captures corrective residual direction.

\subsection{Medical Supervision Calibration}
\label{sec:supervision_calibration}

After obtaining the visual necessity score, ViCoR adjusts the ranking using answer-level supervision compatibility. A visually necessary sample may still provide noisy supervision when its answer is overly verbose, weakly aligned with clinical labels, or contains an excessively dense label set.

We first normalize the supervised answer $y_i$ and extract an answer-label mention list $\mathcal{M}_i=(\mu_{i,1},\ldots,\mu_{i,m_i})$. A fixed alias map links each mention to a canonical clinical vocabulary $\mathcal{V}$; unmapped mentions are retained in $\mathcal{M}_i$ and count toward $m_i$, but not toward $h_i$. Let $m_i$ be the number of parsed mentions and $h_i$ be the number of mentions successfully mapped to $\mathcal{V}$. The alignment feature combines mention-level precision and label support:
\begin{equation}
A_i^{\mathrm{prec}}
=
\frac{h_i}{\max(m_i,1)},
\qquad
A_i^{\mathrm{sup}}
=
\frac{\min(h_i,\tau)}{\tau},
\qquad
A_i
=
\beta_A A_i^{\mathrm{prec}}
+
(1-\beta_A)A_i^{\mathrm{sup}},
\end{equation}
where $\tau$ is the source-specific target answer cardinality and $\beta_A\in[0,1]$ is fixed before evaluation. A larger $A_i$ indicates that the parsed answer mentions are better aligned with the canonical clinical label space.

ViCoR also favors concise classification-style answers. Let $r_a(i)$ and $r_r(i)$ be the percentile ranks of the final-answer length and reasoning length within the training pool, respectively. The directness feature is
\begin{equation}
D_i
=
1-\left(
\beta_D r_a(i)+(1-\beta_D)r_r(i)
\right),
\end{equation}
where $r_a(i),r_r(i)\in[0,1]$ are percentile ranks within the training pool and $\beta_D\in[0,1]$ is fixed before evaluation. Larger $D_i$ corresponds to more concise diagnostic supervision.

Finally, ViCoR uses a soft cardinality factor to penalize overly dense answer sets:
\begin{equation}
P_i
=
\frac{1}{1+\max(m_i-\tau,0)/\eta},
\end{equation}
where $\eta>0$ controls the softness of the penalty. Answers within the target cardinality receive no penalty, while answers exceeding $\tau$ are smoothly down-weighted as the parsed label count increases.
The final ranking score is
\begin{equation}
S_i^{\mathrm{rank}}
=
S_i^{\mathrm{vis}}\cdot B_i,
\qquad
B_i
=
1+\alpha_a A_i+\alpha_d D_i+\alpha_c(P_i-1).
\end{equation}
The calibration factor $B_i$ increases when parsed answer mentions align with the clinical vocabulary and when the answer is concise, while softly down-weighting answers whose parsed label count exceeds the target cardinality. The coefficients $\alpha_a,\alpha_d,\alpha_c\ge0$ are fixed in the selection configuration before evaluation, and chosen so that $B_i$ remains positive over the training pool. This calibration favors visually necessary samples with cleaner and more structured medical supervision.

\subsection{Diversity-Aware Subset Construction}
\label{sec:subset_construction}

Although $S_i^{\mathrm{rank}}$ scores each sample independently, it does not account for redundancy or coverage within the selected subset. Directly selecting the top $K$ samples can therefore overconcentrate the subset in frequent answer patterns or similar visual cases. ViCoR uses $S_i^{\mathrm{rank}}$ to form a high-quality candidate pool, and then applies diversity-aware subset construction to improve coverage and reduce redundancy. The final budget $K$ is fixed by the target selection ratio, which is 20\% in the main experiments.

First, we rank all training examples by $S_i^{\mathrm{rank}}$ and retain an overcomplete candidate pool $\mathcal{C}$ with $|\mathcal{C}|=\min(\gamma K, N)$. Each candidate $d\in\mathcal{C}$ is represented by a lightweight descriptor $\phi_d$, consisting of length features, a hashed answer-pattern feature, and compact residual statistics. The descriptor is used only for redundancy estimation, while $S_i^{\mathrm{rank}}$ remains the quality score.

Second, ViCoR performs diversity-aware re-ranking within $\mathcal{C}$. At each step, it selects
\begin{equation}
d^\ast
=
\arg\max_{d\in\mathcal{C}\setminus\mathcal{S}}
\left[
\lambda \widetilde{S}_d^{\mathrm{rank}}
-
(1-\lambda)
\max_{s\in\mathcal{S}}
\cos(\phi_d,\phi_s)
\right],
\end{equation}

where $\widetilde{S}_d^{\mathrm{rank}}$ is the min-max normalized ranking score within the candidate pool, $\cos(\phi_d,\phi_s)$ measures redundancy between candidate descriptors, and $\lambda\in[0,1]$ controls the quality-redundancy trade-off. The first term favors visually necessary and supervision-compatible samples, while the second term penalizes candidates that are too similar to already selected examples. When $\mathcal{S}$ is empty, the redundancy term is set to zero. The procedure continues until $|\mathcal{S}|=K$. The final subset $\mathcal{S}$ retains high-ranking samples while reducing redundancy across answer patterns and sample structures.

\section{Experiments}

\subsection{Experimental Setup}
\textbf{Training Datasets.} We conduct data selection on three medical multimodal reasoning datasets: GEMeX-ThinkVG~\cite{liu2025gemexrmcotenhancedmedvqadataset}, S-Chain~\cite{le2025s}, and CXR-CounterFact~\cite{yang2025walking}. GEMeX-ThinkVG provides visual grounding annotations, S-Chain emphasizes sequential diagnostic reasoning, and CXR-CounterFact contains counterfactual image-text pairs for feature learning. We further isolate CXR-CounterFact-Multi for evaluating combinatorial long-tail robustness in multi-disease scenarios.

\textbf{Baselines.} All 20\% subset selection methods use the same model backbone, training source, fine-tuning recipe, and evaluation protocol. Full uses 100\% of the training pool, and Random uses a random 20\% subset. We compare ViCoR with GraNd, LESS, Self-Filter, and Adapt-$\infty$, covering gradient-norm sample selection, influence-based data selection, model-based instruction filtering, and dynamic multimodal data selection.

\textbf{Evaluation Benchmarks.} We evaluate the generalization capabilities of the fine-tuned models on four external medical benchmarks. These include OmniMedVQA~\cite{hu2024omnimedvqa} for diverse medical visual question answering, GMAI-MMBench~\cite{ye2024gmai} for broad medical reasoning and capability evaluation, MedBLINK~\cite{11375294} for foundational multimodal visual perception, and DrVD-Bench~\cite{zhou2026drvdbench} for clinical visual reasoning. We report the average score across all sub-tasks for these benchmarks. To assess robustness under overlapping pathologies, we also evaluate on the Multiple Disease Identification task of CheXbench~\cite{chen2024chexagent}, reporting Accuracy and Macro-F1.

\textbf{Implementation and Training Details.} We use the selected data to fine-tune two open-source vision-language models: Qwen3-VL-4B~\cite{bai2025qwen3} and InternVL3.5-4B~\cite{wang2025internvl3}. We use the same fine-tuning recipe for all methods and report the mean performance over three independent runs with different random seeds. Detailed implementation settings are provided in Appendix.

\subsection{Main Results}
\label{sec:main_results}

\textbf{Cross-Benchmark Generalization.}
We begin with the main evaluation under the experimental setup above. Table~\ref{tab:main_single_disease} reports the results when each selection method is applied independently to each training source and the resulting models are evaluated on four external benchmarks. This comparison tests the central question of this work: whether a 20\% subset selected by ViCoR can provide stronger cross-benchmark generalization than full-data fine-tuning or existing data selection strategies. 

\begin{table*}[htbp]
\centering
\caption{Cross-benchmark generalization across three training sources and two backbones. Benchmarks: OmniMedVQA (OMVQA), GMAI-MMBench (GMAI), MedBLINK (MBLK), and DrVD-Bench (DRVD). Avg. is the four-benchmark mean. \textbf{Bold} and \underline{underline} mark best and second-best.}
\label{tab:main_single_disease}

\small
\resizebox{\textwidth}{!}{%
\setlength{\tabcolsep}{2pt}
\renewcommand{\arraystretch}{1.0}
\begin{tabular}{l ccccc ccccc ccccc}

\toprule
\multirow{2}{*}{\raisebox{-0.6ex}{\textbf{Method}}} & \multicolumn{5}{c}{\textbf{GEMeX-ThinkVG}} & \multicolumn{5}{c}{\textbf{S-Chain}} & \multicolumn{5}{c}{\textbf{CXR-CounterFact-Single}} \\
\cmidrule(lr){2-6} \cmidrule(lr){7-11} \cmidrule(l){12-16}
& \textbf{OMVQA} & \textbf{GMAI} & \textbf{MBLK} & \textbf{DRVD} & \textbf{\textit{Avg.}}
& \textbf{OMVQA} & \textbf{GMAI} & \textbf{MBLK} & \textbf{DRVD} & \textbf{\textit{Avg.}}
& \textbf{OMVQA} & \textbf{GMAI} & \textbf{MBLK} & \textbf{DRVD} & \textbf{\textit{Avg.}}\\
\midrule

\rowcolor{cyan!10}
\multicolumn{16}{c}{\textbf{Qwen3-VL-4B}} \\
Full (100\%)   & 82.47 & 45.54 & \textbf{64.30} & 56.75 & \textit{62.27} & 73.66 & \underline{52.35} & 63.37 & \textbf{60.73} & \textit{62.53} & 77.59 & 50.02 & 60.11 & 58.81 & \textit{61.63} \\
Random         & 83.89 & \underline{49.98} & 52.61 & 60.01 & \textit{61.62} & 64.37 & 46.37 & 40.45 & 50.39 & \textit{50.40} & 77.67 & 50.47 & 60.07 & 60.06 & \textit{62.07} \\
GraNd~\cite{paul2021deep}           & 84.93 & 40.12 & 62.59 & \underline{60.30} & \textit{61.99} & 73.28 & 42.82 & 63.29 & 59.62 & \textit{59.75} & 77.08 & 44.79 & 58.38 & 62.42 & \textit{60.67} \\
LESS~\cite{xia2024less}           & 84.46 & 42.58 & 61.57 & 59.05 & \textit{61.92} & \underline{74.33} & 43.31 & 63.76 & 60.33 & \textit{60.43} & \underline{78.29} & 44.79 & 61.03 & 62.07 & \textit{61.55} \\
Adapt-$\infty$~\cite{maharana2025adaptinfty} & 83.88 & 49.69 & 58.30 & 59.97 & \textit{62.96} & 73.86 & 51.78 & \underline{64.54} & 59.99 & \underline{\textit{62.54}} & 74.26 & \underline{51.71} & \underline{61.42} & 60.55 & \textit{61.99} \\
Self-Filter~\cite{chen2024your}    & \underline{85.01} & 49.85 & \underline{63.68} & 59.14 & \underline{\textit{64.42}} & 71.67 & 49.43 & \textbf{65.39} & 59.21 & \textit{61.43} & 76.61 & 51.05 & 58.77 & \underline{62.73} & \underline{\textit{62.29}} \\
\rowcolor{gray!15}
ViCoR (Ours)   & \textbf{85.16} & \textbf{51.56} & \underline{63.68} & \textbf{61.49} & \textbf{\textit{65.47}} & \textbf{75.58} & \textbf{52.66} & 63.68 & \underline{60.34} & \textbf{\textit{63.07}} & \textbf{81.86} & \textbf{53.86} & \textbf{62.67} & \textbf{62.78} & \textbf{\textit{65.29}} \\

\cmidrule{1-16}
\rowcolor{cyan!10}
\multicolumn{16}{c}{\textbf{InternVL3.5-4B}} \\
Full (100\%)   & 87.52 & 27.36 & \textbf{57.83} & 60.79 & \textit{58.38} & \underline{80.61} & 40.49 & 54.79 & \underline{51.48} & \underline{\textit{56.84}} & 80.99 & 39.14 & 38.66 & \textbf{61.84} & \textit{55.16} \\
Random         & 86.18 & 25.64 & 53.70 & 58.18 & \textit{55.93} & 72.58 & 39.75 & 52.02 & 44.92 & \textit{52.32} & 75.32 & 38.28 & 36.08 & 60.57 & \textit{52.56} \\
GraNd~\cite{paul2021deep}          & 87.94 & \underline{36.20} & 50.43 & \textbf{61.98} & \textit{59.14} & 73.18 & 39.26 & \textbf{58.77} & 50.81 & \textit{55.51} & \textbf{84.85} & 40.74 & 38.74 & 60.33 & \textit{56.17} \\
LESS~\cite{xia2024less}           & \underline{88.30} & 33.62 & 57.03 & 61.46 & \underline{\textit{60.10}} & 71.87 & 39.88 & 53.31 & 49.81 & \textit{53.72} & 83.20 & 40.98 & \underline{47.16} & 60.46 & \underline{\textit{57.95}} \\
Adapt-$\infty$~\cite{maharana2025adaptinfty} & \textbf{88.35} & 31.85 & 55.96 & 60.79 & \textit{59.24} & 72.41 & 44.59 & \underline{55.26} & 50.64 & \textit{55.73} & 67.00 & 40.95 & 39.52 & 53.65 & \textit{50.28} \\
Self-Filter~\cite{chen2024your}    & 86.00 & 29.01 & \underline{57.21} & 61.67 & \textit{58.47} & 76.82 & \underline{45.52} & 52.77 & 50.12 & \textit{56.31} & 66.34 & \underline{47.69} & 39.36 & 55.08 & \textit{52.12} \\
\rowcolor{gray!15}
ViCoR (Ours)   & 84.18 & \textbf{39.45} & 55.99 & \underline{61.70} & \textbf{\textit{60.33}} & \textbf{81.69} & \textbf{47.22} & 53.23 & \textbf{54.16} & \textbf{\textit{59.08}} & \underline{83.23} & \textbf{52.13} & \textbf{57.68} & \underline{60.86} & \textbf{\textit{63.48}} \\
\bottomrule
\end{tabular}
}
\end{table*}

The results show that ViCoR delivers the strongest average cross-benchmark generalization. On Qwen3-VL-4B, ViCoR achieves the best average score for all three training sources and leads all four benchmarks when trained on CXR-CounterFact-Single. On InternVL3.5-4B, it again obtains the best average score for all three sources, despite variation in the best method on individual benchmarks. ViCoR also surpasses full-data fine-tuning in every average score while using only 20\% of the data. This suggests that data volume alone is insufficient, and filtering shortcut-compatible supervision can improve cross-benchmark generalization. Appendix~\ref{app:stat_uncertainty} reports paired bootstrap differences, which support the consistency of these gains.

\textbf{Multi-Disease Robustness.}
Multi-disease diagnosis tests whether selected data preserves visually necessary evidence under overlapping pathologies. Unlike single-disease evaluation, the model must identify the correct combination of findings rather than rely on the most salient or frequent condition. This setting is therefore sensitive to shortcut learning: a model can perform well on dominant disease cues while still missing secondary abnormalities. Table~\ref{tab:main_multi_chexbench} evaluates whether ViCoR-selected subsets improve robustness in this combinatorial setting, after fine-tuning on CXR-CounterFact-Multi.

ViCoR obtains the best Accuracy and Macro-F1 on both backbones, outperforming full-data fine-tuning despite using only 20\% of the training data and maintaining clear advantages over most 20\% selection baselines. On Qwen3-VL-4B, ViCoR improves over all subset-selection baselines and shows a particularly clear advantage over LESS and Random. On InternVL3.5-4B, Adapt-$\infty$ remains the closest competitor, while ViCoR still achieves the best overall results. These results suggest that ViCoR-selected subsets retain supervision that remains useful beyond dominant single-disease cues, providing more reliable coverage when multiple findings compete within the same image.

\begin{table*}[htbp]
\vspace{-0.6em}
\centering
\scriptsize 
\renewcommand{\arraystretch}{0.95}

\begin{minipage}[t]{0.48\textwidth}
\vspace{0pt}
\centering
\caption{CheXbench Multiple Disease Identification after fine-tuning on CXR-CounterFact-Multi with selected 20\% subsets.}
\label{tab:main_multi_chexbench}
\vspace{0.6em}
\setlength{\tabcolsep}{3.2pt}
\begin{tabular}{l cc cc}
\toprule
\multirow{2}{*}{\raisebox{-0.8ex}{\textbf{Method}}}
& \multicolumn{2}{c}{\textbf{Qwen3-VL-4B}} 
& \multicolumn{2}{c}{\textbf{InternVL3.5-4B}} \\
\cmidrule(lr){2-3} \cmidrule(lr){4-5}
& \textbf{Accuracy} & \textbf{Macro-F1} 
& \textbf{Accuracy} & \textbf{Macro-F1} \\
\midrule
Full (100\%)    & \underline{44.65} & \underline{44.36} & \underline{51.69} & \underline{41.55} \\
Random          & 40.95 & 40.69 & 27.77 & 27.57 \\
GraNd           & 43.11 & 42.88 & 28.76 & 28.94 \\
LESS            & 40.30 & 39.89 & 32.66 & 32.22 \\
Adapt-$\infty$  & 42.61 & 42.20 & 45.76 & 41.12 \\
Self-Filter     & 43.05 & 42.82 & 25.92 & 24.95 \\
\rowcolor{gray!15}
\textbf{ViCoR (Ours)}  & \textbf{46.84} & \textbf{46.65} & \textbf{52.56} & \textbf{42.59} \\
\bottomrule
\end{tabular}
\end{minipage}
\hfill
\begin{minipage}[t]{0.50\textwidth}
\vspace{0pt}
\centering
\caption{Fine-grained finding-combination discrimination on CheXbench after fine-tuning on CXR-CounterFact-Multi with selected 20\% subsets.}
\label{tab:clinical_stratification}
\vspace{0.6em}
\scriptsize
\setlength{\tabcolsep}{1.8pt}
\renewcommand{\arraystretch}{0.95}
\begin{tabular}{l ccc ccc}
\toprule
\multirow{2}{*}{\raisebox{-0.8ex}{\textbf{Method}}}
& \multicolumn{3}{c}{\textbf{Qwen3-VL-4B}} 
& \multicolumn{3}{c}{\textbf{InternVL3.5-4B}} \\
\cmidrule(lr){2-4} \cmidrule(l){5-7}
& \textbf{C-Acc.} & \textbf{Prob.} & \textbf{NM-Win}
& \textbf{C-Acc.} & \textbf{Prob.} & \textbf{NM-Win} \\
\midrule
Full (100\%)    & \underline{41.49} & 38.80 & \underline{49.34} 
                & \underline{45.42} & \underline{36.37} & \underline{53.71} \\
Random          & 38.75 & 36.08 & 47.17 
                & 26.46 & 26.98 & 38.43 \\
GraNd           & 39.08 & 37.87 & 48.45 
                & 28.12 & 26.76 & 37.71 \\
LESS            & 37.92 & 36.48 & 48.02 
                & 24.73 & 26.13 & 36.84 \\
Adapt-$\infty$  & 41.17 & 38.88 & 49.17 
                & 23.94 & 24.44 & 35.98 \\
Self-Filter     & 40.16 & \underline{38.93} & 49.17 
                & 26.60 & 26.29 & 38.14 \\
\rowcolor{gray!15}
\textbf{ViCoR (Ours)}  & \textbf{41.96} & \textbf{39.76} & \textbf{50.04} 
                & \textbf{49.96} & \textbf{38.90} & \textbf{56.45} \\
\bottomrule
\end{tabular}
\end{minipage}

\vspace{-0.6em}
\end{table*}

\subsection{Fine-Grained Finding-Combination Analysis}
\label{sec:clinical_stratification}

The main CheXbench evaluation measures whether the model selects the correct candidate in a multiple-choice multi-disease task. Each question contains several candidate options, and each option corresponds to a combination of radiographic findings. To better understand model behavior beyond final accuracy, we perform option-level finding-combination scoring on Qwen3-VL-4B and InternVL3.5-4B. For each question, we feed the same image and question to the model, then compute the likelihood assigned to each candidate option label. This avoids confounds from option length, wording, and finding-name tokenization. We report choice accuracy (C-Acc.), correct-option probability (Prob.), and near-miss win rate (NM-Win). Correct-option probability is the average normalized probability of the ground-truth option, while near-miss win rate measures how often the ground-truth option scores higher than the most clinically similar incorrect option.

Table~\ref{tab:clinical_stratification} shows that ViCoR achieves the best result on all three option-level metrics across both backbones. It obtains the highest choice accuracy, assigns the largest probability to the correct candidate, and most frequently ranks the correct candidate above the near-miss distractor. These results indicate that ViCoR improves not only final candidate selection, but also the model's ability to distinguish clinically similar finding combinations.

\subsection{Mechanism Validation}
\label{sec:mechanism_validation}

The main results show performance gains, and we further examine their source. We conduct three mechanism analyses: gradient-level selection statistics test whether ViCoR selects image-dependent training signals, image-level perturbation tests whether the model relies on visual input, and evidence-region perturbation tests whether this reliance concentrates on annotated diagnostic regions.

\vspace{-0.5em}
\paragraph{Gradient-Level Selection Analysis.}

\begin{wraptable}{r}{0.52\textwidth}
\vspace{-1.35em}
\centering
\caption{Gradient-level statistics of selected subsets.}
\vspace{0.5em}
\label{tab:gradient_selection_analysis}
\footnotesize
\setlength{\tabcolsep}{4.5pt} 
\renewcommand{\arraystretch}{0.97}
\resizebox{\linewidth}{!}{%
\begin{tabular}{l ccccc}
\toprule
\textbf{Subset} 
& $\mathbf{E^\Delta}$ 
& $\mathbf{E^s}$ 
& $\mathbf{w^s}$ 
& $\mathbf{w^d}$ 
& $\mathbf{S^{\mathrm{vis}}}$ \\
\midrule
Full (100\%)    & 3.716 & 3.840 & 0.746 & 1.307 & 3.472 \\
Random          & 3.701 & 3.826 & 0.746 & 1.305 & 3.456 \\
GraNd           & 4.739 & 4.622 & 0.724 & 1.338 & 4.352 \\
LESS            & 4.696 & 4.574 & 0.724 & 1.333 & 4.285 \\
Adapt-$\infty$ & 3.255 & 3.444 & 0.756 & 1.304 & 3.078 \\
Self-Filter     & 3.911 & 4.056 & 0.743 & 1.310 & 3.683 \\
\rowcolor{gray!15}
\textbf{ViCoR (Ours)}  & \textbf{7.164} & \textbf{6.876} & \textbf{0.788} & \textbf{1.450} & \textbf{7.856} \\
\bottomrule
\end{tabular}
}
\vspace{-1.0em}
\end{wraptable}

Table~\ref{tab:gradient_selection_analysis} reports subset-level averages to test whether ViCoR selects samples with distinct gradient-level properties before fine-tuning. We conduct this analysis on the GEMeX-ThinkVG training pool with Qwen3-VL-4B. For each training example, we compute residual strength $E_i^\Delta$, shortcut strength $E_i^s$, shortcut dominance weight $w_i^s$, residual-direction weight $w_i^d$, and visual-necessity score $S_i^{\mathrm{vis}}$.

ViCoR-selected samples exhibit the strongest gradient-level statistics of visual necessity among all compared subsets. It obtains the highest residual strength $E^\Delta$ of 7.164, the highest residual-direction weight $w^d$ of 1.450, and the highest visual-necessity score $S^{\mathrm{vis}}$ of 7.856. Random remains close to the full pool across all metrics, while GraNd and LESS increase residual strength but still remain clearly below ViCoR in $S^{\mathrm{vis}}$. These results provide gradient-level evidence that ViCoR preferentially selects samples with stronger image-dependent residual signals before fine-tuning.

\vspace{-0.5em}

\begin{figure*}[htbp]
\centering
\includegraphics[width=0.98\textwidth]{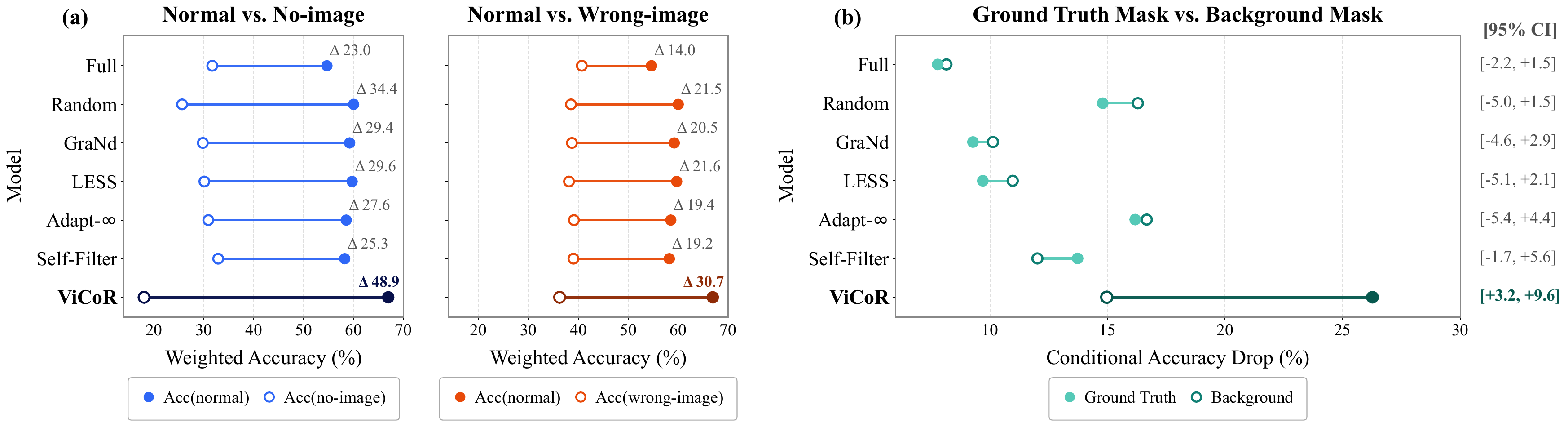}
\vspace{-0.5em}
\caption{
Visual perturbation audits.
(a) Image-level dependence under no-image and wrong-image perturbations.
(b) Region-level dependence under Ground Truth and Background masking, with bootstrap 95\% CIs for the evidence gap.
}
\label{fig:faithfulness_audit}
\vspace{-1.0em}
\end{figure*}

\paragraph{Image-Level Perturbation.}
We conduct an image-level visual perturbation audit to examine whether ViCoR's performance gains are accompanied by stronger visual dependence. From each external evaluation benchmark, we randomly sample 500 examples, resulting in 2,000 examples in total, and evaluate all seven Qwen3-VL-4B models under three input conditions: the original image, a no-image condition that tests dependence on the presence of visual input, and a wrong-image condition that tests dependence on image-content correctness. In this audit, the no-image condition uses a blank input to remove diagnostic visual content while preserving the multimodal input format. In the wrong-image condition, the question is paired with an incorrect image from the same benchmark. Figure~\ref{fig:faithfulness_audit}(a) shows the aggregate accuracy before and after perturbation, where the distance between paired endpoints corresponds to the accuracy drop. When interpreted together with the original accuracy, larger drops are favorable in this audit because they indicate stronger reliance on visual input rather than text-only shortcuts.

ViCoR shows the largest performance degradation under both visual perturbations, while also maintaining the highest aggregate accuracy under the original image condition. Under no-image perturbation, ViCoR drops by 48.9 points, exceeding the strongest competing drop of 34.4 points from Random. Under wrong-image perturbation, ViCoR drops by 30.7 points, also larger than the strongest competing drop of 21.6 points from LESS. This pattern suggests that ViCoR-selected data shifts the model toward image-conditioned prediction rather than question-only shortcut reasoning.

\paragraph{Evidence-Region Perturbation.}
Image-level perturbation tests visual dependence, but not whether correct predictions rely on the annotated evidence region. We conduct an evidence-region audit on the reserved 1,000-example GEMeX grounded set, where evidence boxes enable controlled region masking. Details of the held-out construction are provided in Appendix~\ref{app:dataset_processing}. We first form a shared correct subset of 362 examples correctly answered by every compared model under the original image condition, ensuring that all methods are evaluated on the same examples. All evidence-region drops are then computed on this shared subset for every method. For each example, we compare two matched conditions: a Ground Truth mask that removes the annotated evidence region, and a background mask that removes a non-overlapping region of the same size sampled from the same center distance bin whenever possible. The evidence gap is the additional drop caused by the Ground Truth mask over the Background mask, so a larger positive gap is favorable in this audit because it indicates stronger reliance on the annotated evidence region.

Figure~\ref{fig:faithfulness_audit}(b) shows the strongest evidence-region dependence for ViCoR on the shared-correct subset. For ViCoR, the Ground Truth mask causes a 26.3-point accuracy drop, while the Background mask causes a 16.9-point drop, yielding the largest evidence gap of 9.4 points. Most baselines show near-zero or negative gaps, indicating that masking the annotated evidence region is not consistently more harmful than masking a matched background region. ViCoR is also the only method with a strictly positive 95\% confidence interval, [3.2, 9.6], whereas all baseline intervals cross zero. This provides controlled perturbation evidence that ViCoR's predictions are more sensitive to annotated evidence regions than to arbitrary local regions.

\vspace*{-0.3em}
\raggedbottom

\subsection{Ablation Study}
\label{sec:ablation}
\vspace*{-0.3em}

We conduct a component removal ablation on GEMeX-ThinkVG with Qwen3-VL-4B. All variants use the same 20\% data budget and fine-tuning recipe. Each variant starts from the full ViCoR pipeline and removes one component: counterfactual residual scoring, shortcut weighting, directional correction, residual thresholding, supervision calibration, or diversity-aware reranking. For w/o counterfactual residual, we replace the residual signal with the joint-gradient signal. For w/o diversity reranking, we directly select the top-ranked samples without the diversity-aware construction step.

\begin{wraptable}{r}{0.60\textwidth}
\vspace{-1.35em}
\centering
\caption{Ablation of ViCoR on GEMeX-ThinkVG with Qwen3-VL-4B.}
\vspace{0.4em}
\label{tab:component_ablation}
\setlength{\tabcolsep}{2.2pt}
\renewcommand{\arraystretch}{1.1}
\resizebox{\linewidth}{!}{%
\begin{tabular}{l cccc cc}
\toprule
\multirow{2}{*}{\raisebox{-0.6ex}{\textbf{Configuration}}}
& \multicolumn{4}{c}{\textbf{Single-disease}} 
& \multicolumn{2}{c}{\textbf{Multi-disease}} \\
\cmidrule(lr){2-5} \cmidrule(l){6-7}
& {\small\textbf{OMVQA}} & {\small\textbf{GMAI}} & {\small\textbf{MBLK}} & {\small\textbf{DRVD}}
& {\small\textbf{Acc.}} & {\small\textbf{Macro-F1}} \\
\midrule
w/o Residual      & 82.18 & 42.94 & 56.59 & 58.45 & 43.26 & 43.12 \\
w/o Shortcut      & 81.86 & 41.10 & 57.76 & 59.46 & 44.31 & 44.12 \\
w/o Direction     & 82.18 & 41.10 & 58.69 & 59.86 & 45.13 & 44.91 \\
w/o Threshold     & 80.82 & 50.40 & 58.22 & 59.50 & 45.71 & 45.60 \\
w/o Calibration   & 80.68 & 50.26 & 56.98 & 59.95 & 42.90 & 42.72 \\
w/o Diversity     & 82.44 & 50.35 & 57.83 & 60.13 & 42.25 & 42.18 \\
\rowcolor{gray!15}
\textbf{Full ViCoR} & \textbf{85.16} & \textbf{51.56} & \textbf{63.68} & \textbf{61.49} & \textbf{46.84} & \textbf{46.65} \\
\bottomrule
\end{tabular}
}
\vspace{-0.8em}
\end{wraptable}

Table~\ref{tab:component_ablation} shows that Full ViCoR achieves the best results across all single-disease benchmarks and both multi-disease metrics. Removing counterfactual residual scoring causes broad degradation, confirming the importance of subtracting shortcut-induced updates. Removing shortcut weighting or directional correction further reduces performance, suggesting that residual magnitude alone is insufficient and that ViCoR benefits from modeling shortcut pressure and residual direction. Removing residual thresholding leads to a milder but consistent decline, indicating that the threshold helps suppress negligible residual noise. Supervision calibration and diversity-aware reranking are especially important for multi-disease robustness: removing either component substantially reduces CheXbench performance, with w/o diversity reranking producing the weakest multi-disease results. These findings show that ViCoR’s components provide complementary effects.

\subsection{Data Efficiency}
\label{sec:data_efficiency}

\begin{wrapfigure}{r}{0.49\textwidth}
\vspace{-1.0em}
\centering
\includegraphics[width=\linewidth]{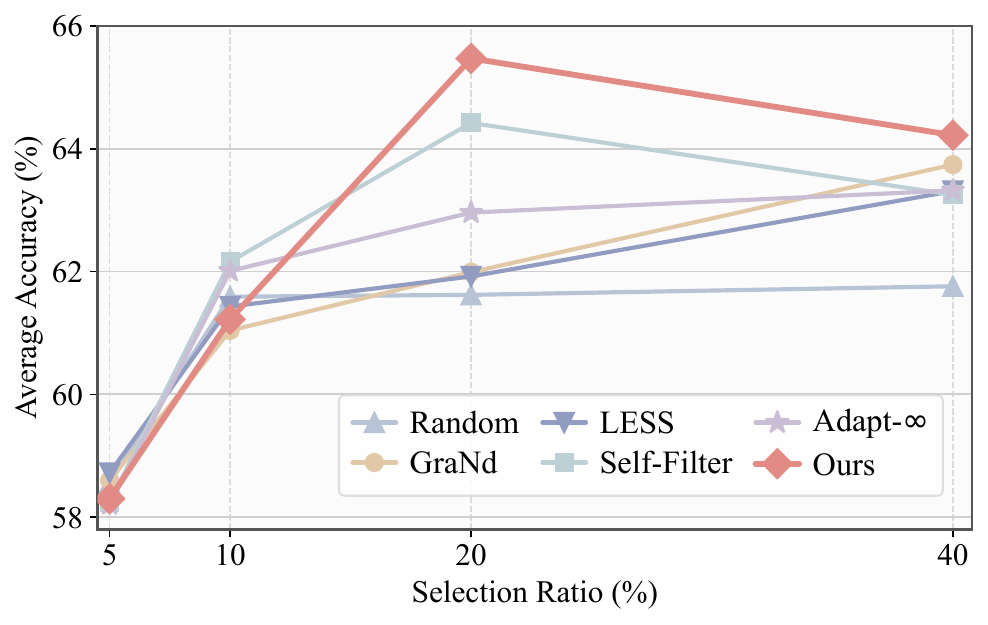}
\vspace{-1.8em}
\caption{Data efficiency under different selection ratios on GEMeX-ThinkVG with Qwen3-VL-4B.}
\label{fig:budget_curve}
\vspace{-1.2em}
\end{wrapfigure}

We further evaluate whether ViCoR remains effective beyond the default 20\% budget. Using GEMeX-ThinkVG with Qwen3-VL-4B, we vary the selection ratio from 5\% to 40\% and compare all subset selection methods under the same training and evaluation protocol. Performance is measured by the average score over the four external benchmarks.

Figure~\ref{fig:budget_curve} shows that all methods perform similarly at very small budgets, where subsets may not preserve clinical coverage. ViCoR gains a clearer advantage as the budget increases, achieving the best average score at 20\% and remaining strongest at 40\%. This supports 20\% as the default budget and suggests that counterfactual residual selection becomes more effective once the subset is large enough to preserve informative supervision.

\enlargethispage{\baselineskip}
\section{Discussion and Conclusion}
\label{sec:conclusion}
This work studies modality imbalance as a distribution-induced optimization pathology in Med-VLM fine-tuning, where language shortcuts dominate updates and starve visually grounded learning. We propose ViCoR, a counterfactual-residual-based data selection method for Med-VLM fine-tuning, which selects a budgeted subset through residual scoring, medical supervision calibration, and diversity-aware construction. Across two vision-language backbones and external medical benchmarks, ViCoR improves cross-benchmark generalization and multi-disease robustness while using substantially less training data. Gradient-level analysis and controlled perturbation audits further link these gains to stronger visual-necessity statistics, increased image-level dependence, and greater sensitivity to annotated evidence regions. Since ViCoR relies on a controlled shortcut environment rather than a perfect causal intervention, visual necessity should be interpreted as an optimization-level proxy rather than a direct causal claim. These results suggest a practical route toward more visually grounded Med-VLM fine-tuning, while clinical deployment still requires prospective validation under real-world distribution shifts.

\section*{Acknowledgments}

This work was supported by the National Natural Science Foundation of China under grant No. 62606277.

{
\bibliographystyle{IEEEtran}
\bibliography{reference}
}

\clearpage
\appendix

\section{Implementation and Reproducibility Protocol}
\label{app:implementation_details}

We summarize the protocol used to reproduce the ViCoR data selection and fine-tuning pipeline. All subset methods are applied to the same training pool under the same data budget, then exported, fine-tuned, and evaluated using the same backbone, training source, fine-tuning recipe, and evaluation protocol. ViCoR is applied as an offline selector before fine-tuning and introduces no inference-time overhead after subset selection. The selected subsets are exported to a multimodal instruction-tuning format, where each sample contains an image, a question, and the supervised answer. The implementation follows fixed scripts, configuration files, and processing utilities for scoring, subset construction, data export, fine-tuning, and evaluation. The key experimental settings required to interpret the results are summarized in the appendix.

\begin{algorithm}[htbp]
\caption{ViCoR Selection and Fine-Tuning Protocol}
\label{alg:vicor_protocol}
\begin{algorithmic}[1]
\Require Training pool $\mathcal{D}$, target budget $K$, base Med-VLM $M$
\Ensure Fine-tuned Med-VLM $M_{\mathcal{S}}$
\State Compute joint and shortcut gradients for each training sample in $\mathcal{D}$
\State Estimate visual necessity using the counterfactual residual score
\State Calibrate the score with supervision compatibility features
\State Construct a diversity-aware subset $\mathcal{S}\subset\mathcal{D}$ with $|\mathcal{S}|=K$
\State Export $\mathcal{S}$ to the multimodal instruction-tuning format
\State Fine-tune $M$ on $\mathcal{S}$ using the same training recipe as other subset methods
\State Evaluate the fine-tuned model on external benchmarks
\State \Return $M_{\mathcal{S}}$
\end{algorithmic}
\end{algorithm}

All selection scores and supervision-compatibility features are computed only from the training pool and fixed clinical terminology. External benchmark labels, test answers, and model predictions are never used for subset construction.

\section{Training Data Processing and Split Construction}
\label{app:dataset_processing}

We describe the data processing procedure used to construct the training pools for ViCoR selection. All processing is performed before subset scoring and fine-tuning. The external evaluation benchmarks are never used during filtering, scoring, calibration, subset construction, or hyperparameter selection.

\paragraph{GEMeX-ThinkVG.}
GEMeX-ThinkVG provides visually grounded medical VQA examples with evidence-box annotations. We first apply a lightweight denoising procedure using a Qwen3-4B filtering model to remove malformed, low-quality, or inconsistent examples. Since GEMeX-ThinkVG contains localized evidence annotations that are useful for the evidence-region perturbation audit, we reserve 1,000 geometry-valid grounded examples before fine-tuning. The remaining 60,000 examples are used as the training pool for data selection and fine-tuning.

\paragraph{S-Chain.}
S-Chain provides structured medical visual reasoning examples. We apply the same Qwen3-4B based denoising procedure to remove invalid or weakly formatted samples. After filtering, the resulting S-Chain training pool contains 10,000 examples. No additional task-specific split is applied to S-Chain.

\paragraph{CXR-CounterFact.}
CXR-CounterFact contains chest X-ray counterfactual image-text examples with heterogeneous task formats. We process this dataset in three steps. First, we apply the same Qwen3-4B based denoising and deduplication procedure used for GEMeX-ThinkVG and S-Chain to remove malformed, duplicated, or low-quality examples. Second, we use Qwen3-4B to classify each remaining example into one of three task types: single-disease, multi-disease, or normal. This classification yields approximately 32,000 single-disease examples, 50,000 multi-disease examples, and 68,000 normal examples before task-format filtering. Third, because part of CXR-CounterFact follows a fill-in-the-blank style that is inconsistent with our decision-oriented VQA fine-tuning setting, we remove these fill-in-the-blank examples from the classified pools.

After task-format filtering, we construct two balanced CXR-CounterFact training sources. For CXR-CounterFact-Single, we combine the retained single-disease examples with randomly sampled normal examples to obtain 40,000 examples in total. For CXR-CounterFact-Multi, we combine the retained multi-disease examples with randomly sampled normal examples and also set the total size to 40,000 examples. This design keeps the single-disease and multi-disease training pools comparable in scale while preserving normal cases as clinically relevant negative examples.

CXR-CounterFact-Single is used for the single-disease CXR setting in the cross-benchmark generalization experiments. CXR-CounterFact-Multi is used as the selection and fine-tuning source for the multi-disease robustness experiments, whose resulting models are evaluated on the CheXbench Multiple Disease Identification task.

\section{Selection and Fine-Tuning Configuration}
\label{app:selection_config}

This section reports the selection and fine-tuning configurations used in the main experiments. All selection hyperparameters are fixed before evaluation. ViCoR allows source-specific target cardinalities, and for the training sources used in this paper we set the target answer cardinality to $\tau=2.0$. External benchmark labels, test answers, and model predictions are never used for scoring, calibration, diversity construction, or configuration selection.

\paragraph{Selection configuration.}
ViCoR uses a residual threshold of $\tau_E=2.5{\times}10^{-5}$, target answer cardinality $\tau=2.0$, shortcut-ratio clipping bounds $\rho_{\min}=0.05$ and $\rho_{\max}=20.0$, and numerical stabilizers $\epsilon=\epsilon_{\cos}=10^{-8}$. For medical supervision calibration, we set the alignment mixture weight to $\beta_A=0.70$, the directness mixture weight to $\beta_D=0.85$, the cardinality softness to $\eta=1.0$, and the calibration weights to $\alpha_a=0.45$, $\alpha_d=0.20$, and $\alpha_c=0.18$. For diversity-aware construction, we use $\lambda=0.85$ and set the candidate-pool multiplier to $\gamma=3$.

\paragraph{Fine-tuning configuration.}
All fine-tuning experiments use LoRA adaptation with bfloat16 precision. The per-device training batch size is $1$, gradient accumulation is $16$, and the effective batch size is approximately $16$. We train for $3$ epochs with learning rate $1{\times}10^{-4}$, cosine learning-rate scheduling, and warmup ratio $0.03$. The LoRA rank is $16$, the LoRA alpha is $32$, and the LoRA dropout is $0.05$. The same fine-tuning configuration is used for full-data fine-tuning and selected-subset fine-tuning under the same backbone and training source. LoRA adapters are applied to \texttt{q\_proj}, \texttt{k\_proj}, \texttt{v\_proj}, \texttt{o\_proj}, \texttt{gate\_proj}, \texttt{up\_proj}, and \texttt{down\_proj}.

\paragraph{Diversity descriptor.}
The descriptor $\phi_d$ used in diversity-aware reranking is used only for redundancy estimation and is separate from the ranking score $S_d^{\mathrm{rank}}$. For each candidate $d$, $\phi_d$ contains continuous length features, a hashed answer-pattern feature, and compact residual statistics. The length features are question length, rationale length, final-answer length, and full sequence length. The residual statistics are $E_d^\Delta$, $\rho_d$, and $\nu_d$, where $\rho_d$ is the raw shortcut-to-residual ratio and $\nu_d$ is the raw directional-novelty term used to form the residual-direction weight. The descriptor does not include label-count features, calibration features $(A_d,D_d,P_d)$, $S_d^{\mathrm{vis}}$, or $S_d^{\mathrm{rank}}$.

\paragraph{Descriptor normalization.}
Before computing cosine similarity in Eq.~(13), continuous dimensions in $\phi_d$ are max-scaled within the candidate pool by dividing each column by its maximum value plus $10^{-8}$. This is max-scaling rather than z-normalization or full min-max normalization. The hashed answer-pattern vector is concatenated with the scaled continuous features, and the final descriptor is row-wise $\ell_2$-normalized. Redundancy is computed as the dot product between normalized descriptors, which is equivalent to cosine similarity.

\paragraph{Hashed answer-pattern feature.}
The hashed answer-pattern feature is built from the normalized final-answer string. We compute an MD5 hash of the normalized answer and map it into a 32-dimensional one-hot bucket vector. No question text, rationale text, or clinical label list is hashed into this descriptor.

\section{Selection Stability under Configuration Perturbations}
\label{app:selection_stability}

We test whether ViCoR selection is sensitive to configuration choices on GEMeX-ThinkVG with Qwen3-VL-4B under the default 20\% budget. For each perturbation, we recompute scores and selected subsets without additional fine-tuning, and compare them with the default configuration. We report Spearman rank correlation over the full pool, Top 20\% overlap before diversity-aware reranking, and Final overlap after the complete subset construction. Calibration perturbations include scaling all calibration weights by $0.5\times$ or $2\times$, and scaling one calibration term by $2\times$ at a time. Scoring and construction perturbations scale the residual threshold by $0.5\times$ or $2\times$, shrink or expand the $\rho$-clipping range, and reduce the diversity coefficient to $0.5\lambda$, since $\lambda$ is constrained to $[0,1]$.

\begin{table}[htbp]
\centering
\caption{Selection stability under configuration perturbations.}
\label{tab:selection_stability_combined}
\small
\setlength{\tabcolsep}{7pt}
\renewcommand{\arraystretch}{1.08}
\begin{tabular}{lccc}
\toprule
\textbf{Perturbation} 
& \textbf{Spearman} 
& \textbf{Top 20\% overlap} 
& \textbf{Final overlap} \\
\midrule
\multicolumn{4}{c}{\textit{Calibration Weights}} \\
\midrule
$0.5\times$ all calibration weights      & 0.9951 & 94.68\%  & 94.80\%  \\
$2\times$ all calibration weights        & 0.9971 & 95.88\%  & 96.05\%  \\
$2\times$ vocabulary weight              & 0.9873 & 90.34\%  & 90.90\%  \\
$2\times$ answer-concision weight        & 0.9989 & 97.98\%  & 98.05\%  \\
$2\times$ label-count weight             & 0.9996 & 99.87\%  & 99.84\%  \\
\midrule
\multicolumn{4}{c}{\textit{Scoring and Construction}} \\
\midrule
$0.5\times$ residual threshold           & 1.000 & 99.58\%  & 99.12\%  \\
$2\times$ residual threshold             & 1.000 & 99.78\%  & 99.55\%  \\
$0.5\times$ $\rho$ clipping range        & 1.000 & 99.61\%  & 99.34\%  \\
$2\times$ $\rho$ clipping range          & 1.000 & 100.00\% & 99.81\%  \\
$0.5\times$ diversity coefficient & 1.000 & 100.00\% & 99.79\% \\
\bottomrule
\end{tabular}
\end{table}

Table~\ref{tab:selection_stability_combined} shows that ViCoR remains stable under these configuration perturbations. Scaling calibration weights or scaling individual calibration terms by $2\times$ preserves high rank consistency, with Spearman correlations above 0.987 and final overlaps above 90\%. Perturbing the residual threshold, shortcut-ratio clipping range, and diversity coefficient changes the selected subset only marginally, with final overlaps remaining at or above 99.12\%. These results indicate that the counterfactual residual score provides the dominant ordering signal, while supervision calibration and diversity-aware construction make stable refinements rather than driving fragile configuration-specific behavior.

\section{Counterfactual Input Robustness}
\label{app:counterfactual_input_robustness}

\begin{wraptable}{r}{0.50\textwidth}
\vspace{-1.0em}
\centering
\caption{Robustness of ViCoR rankings under alternative shortcut controls.}
\vspace{0.4em}
\label{tab:counterfactual_input_robustness}
\footnotesize
\setlength{\tabcolsep}{5pt}
\renewcommand{\arraystretch}{0.98}
\resizebox{\linewidth}{!}{%
\begin{tabular}{lcc}
\toprule
\textbf{Shortcut control} & \textbf{Spearman $\rho$} & \textbf{Top 20\% overlap} \\
\midrule
Blank image    & 0.9675 & 88.31\% \\
Shuffled image & 0.9681 & 88.20\% \\
Wrong image    & 0.9686 & 88.54\% \\
\bottomrule
\end{tabular}
}
\end{wraptable}

ViCoR uses a dataset-level mean image as the default shortcut control to estimate the shortcut gradient. It is computed separately for each training source after backbone-specific resizing and image preprocessing, using a per-channel tensor mean. To examine whether the resulting score is tied to this particular neutral visual control, we recompute the ViCoR scores using three alternative shortcut controls: a blank image, a shuffled image, and a mismatched wrong image.

Table~\ref{tab:counterfactual_input_robustness} reports two stability metrics. Spearman $\rho$ denotes the rank correlation between each alternative shortcut-control ranking and the default mean-image ranking, while Top 20\% overlap measures the overlap between the corresponding selected subsets. The alternative controls yield highly consistent rankings and overlapping selected subsets, suggesting that ViCoR's ranking is stable across different non-diagnostic visual controls and is unlikely to be an artifact of a particular shortcut input.

\section{Supplementary Proxy Signal Analysis}
\label{app:proxy_signal_analysis}
\vspace*{-0.3em}

To complement the main component ablation, we compare ViCoR with proxy-based selectors and a reduced-pipeline baseline under the same 20\% data budget and Qwen3-VL-4B fine-tuning recipe. This analysis tests whether difficulty, raw gradient magnitude, residual magnitude, or supervision calibration with diversity alone can recover the benefit of the full ViCoR selector. Loss Gap ranks samples by the scalar loss difference between original-image and no-image inputs. Joint Gradient ranks samples by multimodal gradient magnitude without removing shortcut-induced updates. Residual Norm ranks samples only by the magnitude of the counterfactual residual. Calib.+Div. only keeps supervision calibration and diversity-aware construction without counterfactual residual scoring.

\begin{table}[htbp]
\centering
\caption{Supplementary proxy-signal and reduced-pipeline comparison on GEMeX-ThinkVG with Qwen3-VL-4B.}
\label{tab:app_proxy_signal_analysis}
\scriptsize
\setlength{\tabcolsep}{3.8pt}
\renewcommand{\arraystretch}{0.98}
\resizebox{0.78\textwidth}{!}{%
\begin{tabular}{l cccc cc}
\toprule
\multirow{2}{*}{\raisebox{-0.6ex}{\textbf{Configuration}}}
& \multicolumn{4}{c}{\textbf{Single-disease}} 
& \multicolumn{2}{c}{\textbf{Multi-disease}} \\
\cmidrule(lr){2-5} \cmidrule(l){6-7}
& \textbf{OMVQA} & \textbf{GMAI} & \textbf{MBLK} & \textbf{DRVD}
& \textbf{Accuracy} & \textbf{Macro-F1} \\
\midrule
w/ Loss Gap              & 83.22 & 40.98 & 59.17 & 60.86 & 41.02 & 40.57 \\
w/ Joint Gradient        & 82.17 & 43.19 & 59.70 & 58.76 & 41.67 & 41.63 \\
w/ Residual Norm         & 82.35 & 41.96 & 60.25 & 59.01 & 44.92 & 44.82 \\
Calib.+Div. only         & 82.02 & 41.28 & 58.48 & 58.66 & 40.82 & 41.27 \\
\rowcolor{gray!15}
\textbf{Full ViCoR}      & \textbf{85.16} & \textbf{51.56} & \textbf{63.68} & \textbf{61.49} & \textbf{46.84} & \textbf{46.65} \\
\bottomrule
\end{tabular}
}
\end{table}

Table~\ref{tab:app_proxy_signal_analysis} shows that neither proxy signals nor the reduced pipeline recover the full benefit of ViCoR. Loss Gap and Joint Gradient remain clearly below the full selector, indicating that scalar perturbation difficulty and raw gradient magnitude are insufficient for visually necessary data selection in this setting. Residual Norm improves multi-disease performance over these proxy signals, suggesting that counterfactual subtraction helps isolate image-dependent updates. Calib.+Div. only also remains below Full ViCoR, showing that cleaner supervision and diversity-aware construction alone do not explain the gains. These results support the design choice of using residual magnitude as a core signal while further incorporating shortcut pressure, residual direction, supervision calibration, and diversity-aware subset construction.

\section{Visual Dependence Audit before Fine-Tuning}
\label{app:auto_visual_necessity_audit}

We examine whether ViCoR's visual necessity score aligns with image-dependent behavior before subset fine-tuning. Using the base Qwen3-VL-4B model on GEMeX-ThinkVG training samples, we compute the final decision-token loss under three conditions: the original image, a no-image perturbation that removes diagnostic visual content while preserving the multimodal input format, and a wrong-image perturbation that pairs the question with a mismatched image from the same training pool. For each sample, we define the automatic visual-dependence target as the larger loss increase caused by the no-image or wrong-image perturbation. A larger value means that removing or corrupting visual information makes the gold decision less likely under the base model.

\begin{table}[htbp]
\centering
\caption{Pre-fine-tuning visual-dependence audit on GEMeX-ThinkVG with the base Qwen3-VL-4B.}
\label{tab:auto_visual_necessity_audit}
\small
\setlength{\tabcolsep}{5.5pt}
\renewcommand{\arraystretch}{1.08}
\begin{tabular}{lcccc}
\toprule
\textbf{Score} 
& \textbf{Top 20\% Mean}
& \textbf{AUC}
& \textbf{Precision@20\%}
& \textbf{$\Delta$ z-Score} \\
\midrule
Decision-token loss       & 0.603 & 0.566 & 38.61 & 0.258 \\
Joint gradient norm       & 0.655 & 0.582 & 39.47 & 0.230 \\
GraNd score               & 0.612 & 0.576 & 38.18 & 0.234 \\
Residual norm $E^\Delta$  & 0.689 & 0.604 & 40.24 & 0.303 \\
\rowcolor{gray!15}
ViCoR $S^{\mathrm{vis}}$  & \textbf{0.693} & \textbf{0.609} & \textbf{41.94} & \textbf{0.312} \\
\bottomrule
\end{tabular}
\end{table}

We compare ViCoR's visual necessity score with four alternative scoring rules: decision-token loss, joint gradient norm, GraNd score, and residual norm. We report four metrics. Top 20\% Mean reports the average visual-dependence target among the top 20\% samples ranked by each score. AUC and Precision@20\% are computed by treating the top 30\% of the visual-dependence target as pseudo-positive samples and the bottom 30\% as pseudo-negative samples. $\Delta$ z-Score denotes the mean standardized score difference between pseudo-positive and pseudo-negative samples.

Table~\ref{tab:auto_visual_necessity_audit} shows that ViCoR's visual necessity score has the strongest alignment with automatic visual-dependence signals among the compared scoring rules. It achieves the highest Top 20\% Mean, AUC, Precision@20\%, and $\Delta$ z-Score. Residual norm $E^\Delta$ is the closest alternative, which is expected because it directly measures the magnitude of the image-dependent residual. However, the full ViCoR score further improves over residual norm by incorporating shortcut pressure and residual direction, suggesting that these additional terms help refine the residual signal rather than dilute it.

These results indicate that ViCoR's visual necessity score retains sample-level alignment with automatic visual-dependence behavior before fine-tuning. The audit also complements the downstream ablations: residual magnitude alone is a useful visual-dependence signal, but the full ViCoR score provides a stronger selection criterion for constructing the final fine-tuning subset.

\section{Selected Data Characterization}
\label{app:selected_data_characterization}

We first characterize the supervision quality and clinical label coverage of the selected subsets on GEMeX-ThinkVG. All statistics are computed before fine-tuning using training-pool metadata, calibration features, canonical clinical label mappings, and selected sample identifiers. We compare the full pool, Random, four selection baselines, and ViCoR, with all subset selection methods using the same 20\% budget.

\begin{table}[htbp]
\centering
\caption{Supervision quality and label coverage on GEMeX-ThinkVG.}
\label{tab:selected_data_supervision_coverage}
\scriptsize
\setlength{\tabcolsep}{4pt}
\renewcommand{\arraystretch}{1.12}
\resizebox{\textwidth}{!}{%
\begin{tabular}{lcccccccc}
\toprule
\textbf{Subset}
& \textbf{Vocab. Match}
& \textbf{Answer Conc.}
& \textbf{Label Ctrl.}
& \textbf{Answer Len.}
& \textbf{Reason Len.}
& \textbf{Label Ent.}
& \textbf{Rare Cov.}
& \textbf{Top Conc.} \\
\midrule
Full pool              & 0.4220 & 0.5019 & 0.4846 & 4.6191 & 112.1705 & 2.0530 & 1.0000 & 0.3569 \\
Random                 & 0.4254 & 0.5016 & 0.4922 & 4.6037 & 112.1304 & 2.0544 & 1.0000 & 0.3476 \\
GraNd                  & 0.5334 & 0.6289 & 0.5746 & 3.9505 & 108.0859 & 1.6018 & 0.7500 & 0.5869 \\
LESS                   & 0.5453 & 0.6355 & 0.5841 & 3.9221 & 108.3751 & 1.6146 & 0.7500 & 0.5781 \\
Adapt-$\infty$        & 0.4192 & 0.3869 & 0.5212 & 5.1817 & 117.0334 & 2.1770 & 1.0000 & 0.2592 \\
Self-Filter            & 0.3617 & 0.5496 & 0.3957 & 4.4513 & 109.5884 & 2.0281 & 0.7500 & 0.3896 \\
\rowcolor{gray!15}
\textbf{ViCoR (Ours)}         & \textbf{0.5502} & \textbf{0.8602} & 0.5145 & \textbf{2.5690} & 110.2629 & \textbf{2.4028} & \textbf{1.0000} & 0.2683 \\
\bottomrule
\end{tabular}
}
\end{table}

Table~\ref{tab:selected_data_supervision_coverage} shows that ViCoR-selected samples have a strong supervision-quality signature. ViCoR obtains the highest vocabulary-match and answer-concision scores, indicating that its selected answers are both better mapped to canonical clinical labels and more directly expressed as diagnostic supervision. It also has the shortest average answer length, suggesting that ViCoR preferentially selects concise finding-oriented answers rather than verbose or weakly structured responses. At the same time, although the calibration uses a fixed alias map, ViCoR achieves the highest label entropy and full rare-label coverage while maintaining a low top-label concentration, suggesting that calibration does not collapse selection toward only frequent canonical labels.
We next characterize the visual evidence geometry of the selected subsets using GEMeX grounding annotations. Evidence boxes are defined in the 336$\times$336 GEMeX coordinate frame. Area ratio measures the fraction of the image covered by evidence boxes, small evidence is defined by the lowest-quartile full-pool area ratio, peripheral evidence is defined by the highest-quartile full-pool center distance, and hard evidence is their union.

\begin{table}[htbp]
\centering
\caption{Visual evidence geometry on GEMeX-ThinkVG.}
\label{tab:selected_data_visual_evidence}
\scriptsize
\setlength{\tabcolsep}{5pt}
\renewcommand{\arraystretch}{1.12}
\resizebox{\textwidth}{!}{%
\begin{tabular}{lcccccc}
\toprule
\textbf{Subset}
& \textbf{Area Ratio}
& \textbf{Small Rate}
& \textbf{Periph. Dist.}
& \textbf{Periph. Rate}
& \textbf{Hard Rate}
& \textbf{Avg. Boxes} \\
\midrule
Full pool              & 0.2475 & 0.2501 & 0.1800 & 0.2500 & 0.3885 & 1.0585 \\
Random                 & 0.2464 & 0.2427 & 0.1705 & 0.2482 & 0.3810 & 1.0574 \\
GraNd                  & 0.2574 & 0.2276 & 0.1631 & 0.2052 & 0.3454 & 1.0269 \\
LESS                   & 0.2556 & 0.2428 & 0.1661 & 0.2116 & 0.3602 & 1.0254 \\
Adapt-$\infty$        & 0.2384 & 0.2569 & 0.1897 & 0.2793 & 0.4131 & 1.0501 \\
Self-Filter            & 0.2386 & 0.2553 & 0.1862 & 0.2666 & 0.4084 & 1.0353 \\
\rowcolor{gray!15}
\textbf{ViCoR (Ours)}         & \textbf{0.2611} & \textbf{0.2676} & \textbf{0.1941} & \textbf{0.2859} & \textbf{0.4275} & \textbf{1.0685} \\
\bottomrule
\end{tabular}
}
\end{table}

Table~\ref{tab:selected_data_visual_evidence} shows that ViCoR selects visually distinctive evidence cases. ViCoR has the highest evidence area ratio, indicating broader annotated evidence coverage. At the same time, ViCoR achieves the highest small-evidence rate, peripheral distance, peripheral evidence rate, and hard-evidence rate, indicating that it retains challenging evidence patterns rather than only large or central findings. Together with Table~\ref{tab:selected_data_supervision_coverage}, these results suggest that ViCoR selects training examples with cleaner medical supervision, broader label coverage, and more visually distinctive evidence patterns.

\section{Statistical Uncertainty}
\label{app:stat_uncertainty}

The main tables report the mean performance over three independent runs with different random seeds. This appendix reports bootstrap 95\% confidence intervals for the main aggregate metrics and paired bootstrap differences for the main cross-benchmark average. For cross-benchmark generalization, Avg. is computed as the unweighted mean of OMVQA, GMAI-MMBench, MedBLINK, and DrVD-Bench, with confidence intervals estimated by benchmark-stratified bootstrap over evaluation examples. For paired differences, each bootstrap round resamples examples within each benchmark, computes the ViCoR-minus-baseline difference for each benchmark, and then averages the four benchmark-level differences. For CheXbench Multiple Disease Identification, confidence intervals are computed by bootstrapping evaluation examples for Accuracy and Macro-F1. These intervals quantify evaluation-sample uncertainty and complement the seed-averaged results in the main tables.

\begin{table*}[htbp]
\centering
\scriptsize
\caption{Cross-benchmark average score with 95\% CI.}
\label{tab:avg_ci}
\setlength{\tabcolsep}{4.5pt}
\renewcommand{\arraystretch}{1.12}
\resizebox{\textwidth}{!}{%
\begin{tabular}{llccc}
\toprule
\textbf{Backbone} & \textbf{Method} 
& \textbf{GEMeX-ThinkVG} 
& \textbf{S-Chain} 
& \textbf{CXR-CounterFact-Single} \\
\midrule
\multirow{7}{*}{Qwen3-VL-4B}
& Full 
& 62.27 [61.13, 63.34] 
& 62.53 [61.39, 63.62] 
& 61.63 [60.56, 62.74] \\
& Random 
& 61.62 [60.41, 62.75] 
& 50.40 [49.27, 51.50] 
& 62.07 [60.94, 63.25] \\
& GraNd 
& 61.99 [60.86, 63.12] 
& 59.75 [58.67, 60.77] 
& 60.67 [59.52, 61.82] \\
& LESS 
& 61.92 [60.80, 62.98] 
& 60.43 [59.28, 61.47] 
& 61.55 [60.48, 62.63] \\
& Adapt-$\infty$ 
& 62.96 [61.76, 64.06] 
& 62.54 [61.44, 63.62] 
& 61.99 [60.85, 63.10] \\
& Self-Filter 
& 64.42 [63.31, 65.50] 
& 61.43 [60.36, 62.49] 
& 62.29 [61.10, 63.42] \\
\rowcolor{gray!15}
& \textbf{ViCoR (Ours)} 
& \textbf{65.47 [64.38, 66.60]} 
& \textbf{63.07 [62.06, 64.07]} 
& \textbf{65.29 [64.16, 66.53]} \\
\midrule
\multirow{7}{*}{InternVL3.5-4B}
& Full 
& 58.38 [57.25, 59.51] 
& 56.84 [55.68, 57.98] 
& 55.16 [54.06, 56.27] \\
& Random 
& 55.93 [54.86, 57.01] 
& 52.32 [51.31, 53.31] 
& 52.56 [51.68, 53.50] \\
& GraNd 
& 59.14 [58.05, 60.20] 
& 55.51 [54.41, 56.63] 
& 56.17 [55.12, 57.28] \\
& LESS 
& 60.10 [59.02, 61.24] 
& 53.72 [52.57, 54.87] 
& 57.95 [56.84, 59.05] \\
& Adapt-$\infty$ 
& 59.24 [58.15, 60.35] 
& 55.73 [54.63, 56.83] 
& 50.28 [49.21, 51.38] \\
& Self-Filter 
& 58.47 [57.41, 59.56] 
& 56.31 [55.17, 57.46] 
& 52.12 [50.97, 53.27] \\
\rowcolor{gray!15}
& \textbf{ViCoR (Ours)} 
& \textbf{60.33 [59.22, 61.47]} 
& \textbf{59.08 [58.01, 60.15]} 
& \textbf{63.48 [62.35, 64.67]} \\
\bottomrule
\end{tabular}
}
\end{table*}

\begin{table}[ht]
\centering
\small
\caption{CheXbench Multiple Disease Identification with 95\% confidence intervals.}
\label{tab:chex_ci}
\setlength{\tabcolsep}{5pt}
\renewcommand{\arraystretch}{1.12}
\begin{tabular}{llcc}
\toprule
\textbf{Backbone} & \textbf{Method} & \textbf{Accuracy [95\% CI]} & \textbf{Macro-F1 [95\% CI]} \\
\midrule
\multirow{7}{*}{Qwen3-VL-4B}
& Full & 44.65 [42.05, 47.46] & 44.36 [41.77, 47.08] \\
& Random & 40.95 [38.36, 43.69] & 40.69 [38.01, 43.43] \\
& GraNd & 43.11 [40.52, 45.71] & 42.88 [40.19, 45.46] \\
& LESS & 40.30 [37.78, 42.97] & 39.89 [37.39, 42.60] \\
& Adapt-$\infty$ & 42.61 [39.94, 45.06] & 42.20 [39.40, 44.67] \\
& Self-Filter & 43.05 [40.45, 45.72] & 42.82 [40.10, 45.43] \\
\rowcolor{gray!15}
& \textbf{ViCoR (Ours)} & \textbf{46.84 [44.28, 49.40]} & \textbf{46.65 [44.21, 49.34]} \\
\midrule
\multirow{7}{*}{InternVL3.5-4B}
& Full & 51.69 [49.24, 54.29] & 41.55 [39.54, 43.71] \\
& Random & 27.77 [25.31, 30.14] & 27.57 [25.25, 29.83] \\
& GraNd & 28.76 [26.23, 31.14] & 28.94 [26.63, 31.14] \\
& LESS & 32.66 [30.21, 34.97] & 32.22 [30.02, 34.18] \\
& Adapt-$\infty$ & 45.76 [43.16, 48.28] & 41.12 [38.93, 43.27] \\
& Self-Filter & 25.92 [23.72, 28.28] & 24.95 [22.66, 26.97] \\
\rowcolor{gray!15}
& \textbf{ViCoR (Ours)} & \textbf{52.56 [49.75, 55.30]} & \textbf{42.59 [40.33, 44.85]} \\
\bottomrule
\end{tabular}
\end{table}

\begin{table*}[htbp]
\centering
\small
\caption{Paired bootstrap differences between ViCoR and competing methods. }
\label{tab:paired_bootstrap_vicor}
\setlength{\tabcolsep}{4pt}
\renewcommand{\arraystretch}{1.08}
\resizebox{\textwidth}{!}{%
\begin{tabular}{llcc}
\toprule
\textbf{Backbone} & \textbf{Training Source} & \textbf{ViCoR minus Full} & \textbf{ViCoR minus Best 20\% Baseline} \\
\midrule
Qwen3-VL-4B & GEMeX
& $+3.20$ [$+2.45$, $+3.95$]
& $+1.05$ [$+0.35$, $+1.75$] \\
Qwen3-VL-4B & S-Chain
& $+0.54$ [$-0.05$, $+1.13$]
& $+0.53$ [$-0.06$, $+1.10$] \\
Qwen3-VL-4B & CXR Single
& $+3.66$ [$+2.85$, $+4.45$]
& $+3.00$ [$+2.25$, $+3.75$] \\
InternVL3.5-4B & GEMeX
& $+1.95$ [$+1.15$, $+2.75$]
& $+0.23$ [$-0.45$, $+0.90$] \\
InternVL3.5-4B & S-Chain
& $+2.24$ [$+1.45$, $+3.05$]
& $+2.77$ [$+1.95$, $+3.60$] \\
InternVL3.5-4B & CXR Single
& $+8.32$ [$+7.35$, $+9.25$]
& $+5.53$ [$+4.65$, $+6.40$] \\
\bottomrule
\end{tabular}
}
\end{table*}

\paragraph{Interpretation.}
ViCoR obtains the highest point-estimate cross-benchmark average in all six backbone-source settings. The largest margins appear on CXR-CounterFact-Single for both backbones, while the closest comparisons occur on Qwen3-VL-4B with S-Chain and InternVL3.5-4B with GEMeX-ThinkVG. The paired difference analysis further distinguishes large improvements from close-call comparisons: ViCoR shows consistently positive point-estimate differences over both full-data fine-tuning and the strongest 20\% subset baseline. The paired intervals are clearly positive in four of the six comparisons against the strongest 20\% baseline, while the two closest settings include zero and are therefore interpreted as positive point-estimate trends. On CheXbench Multiple Disease Identification, ViCoR also achieves the highest point estimates for both Accuracy and Macro-F1 on both backbones. These results support a consistent data-efficient advantage.

\section{Computational Cost}
\label{app:computational_cost}

\begin{wrapfigure}{r}{0.53\textwidth}
\vspace{-0.5em}
\centering
\includegraphics[width=\linewidth]{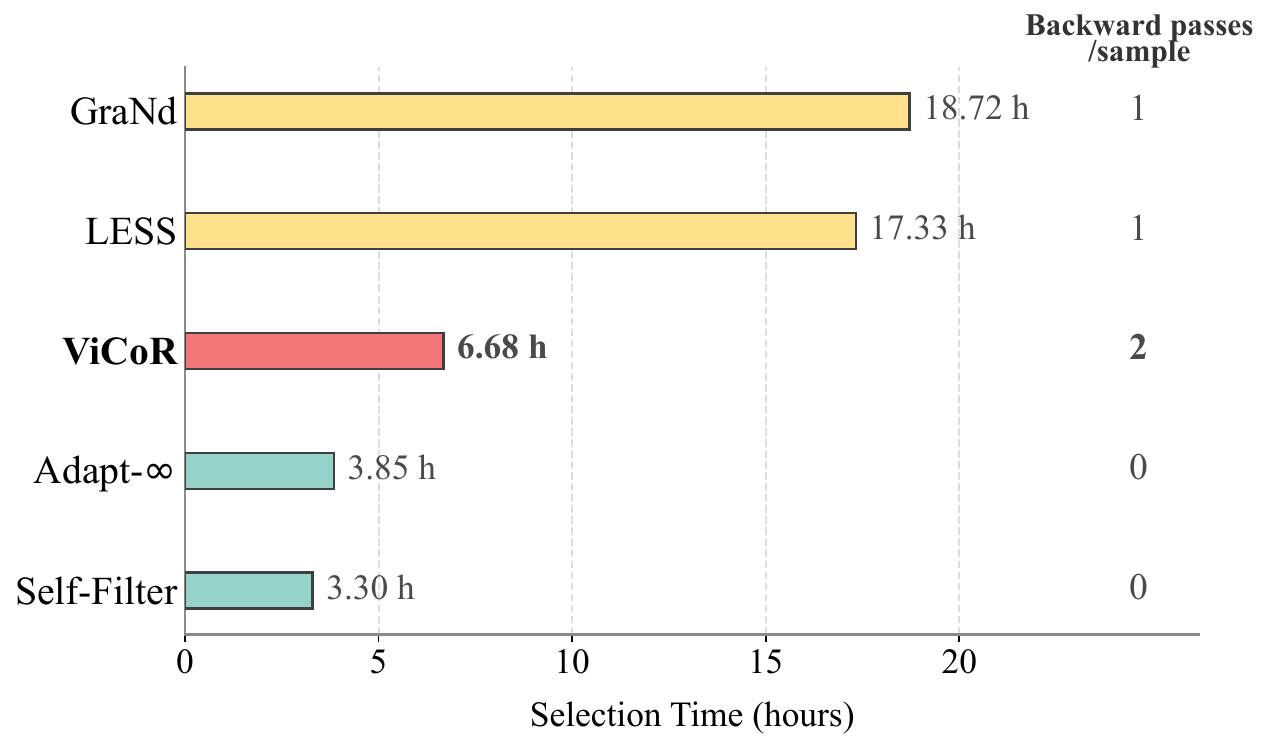}
\vspace{-1.2em}
\caption{Offline selection time on GEMeX-ThinkVG with Qwen3-VL-4B. Right-side labels show backward passes per sample.}
\label{fig:selection_cost}
\vspace{-0.8em}
\end{wrapfigure}

We summarize computational cost using two complementary quantities: selection time and backward passes per sample. Selection time measures the offline wall-clock cost before fine-tuning, while backward passes per sample indicates whether a selector requires gradient computation during scoring. All selection times are measured on NVIDIA RTX 5880 48GB GPUs under the same scoring configuration for GEMeX-ThinkVG with Qwen3-VL-4B.

Figure~\ref{fig:selection_cost} shows that ViCoR requires two backward passes per sample, corresponding to the joint and shortcut gradients used for counterfactual residual scoring. Backward-pass count reflects the algorithmic gradient requirement, but does not directly determine wall-clock time because different selectors use different gradient scopes, scoring targets, feature caching strategies, and aggregation procedures. In particular, GraNd and LESS require storing or aggregating per-sample gradient features over broader scoring targets, whereas ViCoR restricts gradient computation to decision-token losses over the trainable parameter subset and reuses non-visual preprocessing across the joint and shortcut scoring passes whenever possible. In practice, ViCoR takes 6.68 hours, substantially less than GraNd and LESS, while remaining an offline selector with no inference-time overhead after fine-tuning.

\section{Limitations}
\label{app:limitations}

ViCoR estimates visual necessity as an optimization-level proxy rather than a direct causal effect. The mean-image shortcut control removes sample-specific visual content, but it cannot cover all possible shortcut environments or eliminate every nonvisual correlation. Thus, the counterfactual residual should be interpreted as a practical estimate of image-dependent training signal. Our perturbation audits provide controlled evidence of stronger image dependence and annotated evidence region sensitivity, but they remain limited by benchmark coverage, available region annotations, and the shared-correct subset used for masking. ViCoR also incurs offline scoring cost because each candidate requires gradient computation under joint and shortcut inputs. Although this cost is incurred only once before fine-tuning, larger multimodal corpora may require approximate or batched scoring. Finally, clinical deployment requires prospective validation under real-world distribution shifts and institution-specific data governance constraints.

\clearpage
\section*{NeurIPS Paper Checklist}

\begin{enumerate}

\item \textbf{Claims}

\textbf{Question:} Do the main claims made in the abstract and introduction accurately reflect the paper's contributions and scope?

\textbf{Answer:} \answerYes{}

\textbf{Justification:}
\begin{itemize}
    \item The claims describe ViCoR as an offline counterfactual-residual data selection framework for medical vision-language fine-tuning.
    \item The claimed benefits are limited to data-efficient fine-tuning, cross-benchmark generalization, multi-disease robustness, and controlled perturbation evidence of stronger image dependence and annotated-region sensitivity.
    \item These claims are supported by the main results, component ablations, gradient-level analyses, image-level perturbation audits, evidence-region perturbation audits, and statistical uncertainty analyses reported in the paper.
\end{itemize}

\item \textbf{Limitations}

\textbf{Question:} Does the paper discuss the limitations of the work performed by the authors?

\textbf{Answer:} \answerYes{}

\textbf{Justification:}
\begin{itemize}
    \item The paper discusses that visual necessity is estimated through an optimization-level proxy rather than a direct causal intervention.
    \item It notes that the shortcut-control environment cannot cover all possible shortcut conditions.
    \item It states that region-level audits require localized evidence annotations and are limited by benchmark coverage and the shared-correct evaluation subset used for controlled masking.
    \item It also discusses the offline scoring cost introduced by ViCoR and the need for prospective validation before clinical deployment.
\end{itemize}

\item \textbf{Theory assumptions and proofs}

\textbf{Question:} For each theoretical result, does the paper provide the full set of assumptions and a complete and correct proof?

\textbf{Answer:} \answerNA{}

\textbf{Justification:}
\begin{itemize}
    \item The paper does not claim formal theoretical guarantees or present proof-based results.
    \item It defines an optimization-based scoring objective, supervision calibration terms, and a diversity-aware subset-construction procedure.
    \item These components are validated empirically through ablations, robustness analyses, and perturbation audits.
\end{itemize}

\item \textbf{Experimental result reproducibility}

\textbf{Question:} Does the paper fully disclose all the information needed to reproduce the main experimental results of the paper to the extent that it affects the main claims and conclusions of the paper?

\textbf{Answer:} \answerYes{}

\textbf{Justification:}
\begin{itemize}
    \item The paper specifies the training sources, evaluation benchmarks, model backbones, baselines, data budget, scoring loss, shortcut-control design, subset-construction protocol, perturbation settings, and seed-averaged reporting.
    \item The appendix provides additional implementation details for data construction, scoring, selection, fine-tuning, evaluation, perturbation audits, uncertainty estimation, and compute cost.
    \item External datasets, benchmarks, and model checkpoints are not redistributed and must be obtained from their original providers under the corresponding licenses and terms of use.
\end{itemize}

\item \textbf{Open access to data and code}

\textbf{Question:} Does the paper provide open access to the data and code, with sufficient instructions to faithfully reproduce the main experimental results?

\textbf{Answer:} \answerNA{}

\textbf{Justification:}
\begin{itemize}
    \item The submission does not include source code, model checkpoints, or redistributed medical datasets.
    \item The experiments use existing public or access-controlled datasets, benchmarks, and open-source model backbones, which are cited in the paper and must be obtained from their original providers.
    \item We plan to release code and reproducibility materials after the review process.
\end{itemize}

\item \textbf{Experimental setting/details}

\textbf{Question:} Does the paper specify all the training and test details necessary to understand the results?

\textbf{Answer:} \answerYes{}

\textbf{Justification:}
\begin{itemize}
    \item The paper specifies the training datasets, evaluation benchmarks, backbones, selection budget, baseline methods, scoring setup, perturbation protocols, and seed-averaged reporting.
    \item It describes the four-benchmark average, CheXbench Multiple Disease Identification metrics, option-level finding-combination metrics, and the construction of image-level and region-level audits.
    \item The appendix further reports implementation settings for ViCoR scoring, selection hyperparameters, fine-tuning configuration, shortcut-control robustness, uncertainty estimation, and compute cost.
\end{itemize}

\item \textbf{Experiment statistical significance}

\textbf{Question:} Does the paper report error bars suitably and correctly defined or other appropriate information about statistical significance?

\textbf{Answer:} \answerYes{}

\textbf{Justification:}
\begin{itemize}
    \item The main results report mean performance over three independent runs.
    \item The appendix reports bootstrap 95\% confidence intervals for cross-benchmark averages, paired bootstrap differences between ViCoR and competing methods, CheXbench confidence intervals for Accuracy and Macro-F1, and evidence-gap confidence intervals for region-level perturbation.
\end{itemize}

\item \textbf{Experiments compute resources}

\textbf{Question:} Does the paper provide sufficient information on the compute resources needed to reproduce the experiments?

\textbf{Answer:} \answerYes{}

\textbf{Justification:}
\begin{itemize}
    \item The appendix reports the offline scoring cost of ViCoR, including the number of backward passes per sample and measured selection time under the stated GPU configuration.
    \item The paper states that ViCoR introduces no inference-time overhead after subset selection.
    \item Fine-tuning and evaluation use the same recipe across compared methods, and implementation details needed to understand the reported compute setting are provided in the appendix.
\end{itemize}

\item \textbf{Code of ethics}

\textbf{Question:} Does the research conducted in the paper conform to the NeurIPS Code of Ethics?

\textbf{Answer:} \answerYes{}

\textbf{Justification:}
\begin{itemize}
    \item The work uses existing medical datasets and benchmarks and does not involve new patient data collection, participant recruitment, or clinical intervention.
    \item The paper avoids clinical deployment claims and discusses limitations related to shortcut controls, benchmark-based evaluation, evidence annotation requirements, and prospective clinical validation.
    \item The paper does not redistribute restricted raw medical data or patient-level records.
\end{itemize}

\item \textbf{Broader impacts}

\textbf{Question:} Does the paper discuss both potential positive and negative societal impacts?

\textbf{Answer:} \answerYes{}

\textbf{Justification:}
\begin{itemize}
    \item The paper discusses the potential benefit of reducing shortcut reliance and improving visually grounded fine-tuning for medical VLMs.
    \item It also acknowledges risks from benchmark overinterpretation, imperfect causal interpretation of shortcut controls, dependence on localized evidence annotations for some audits, and clinical distribution shifts.
    \item The paper states that prospective validation is required before any clinical deployment.
\end{itemize}

\item \textbf{Safeguards}

\textbf{Question:} Does the paper describe safeguards for responsible release of high-risk assets?

\textbf{Answer:} \answerNA{}

\textbf{Justification:}
\begin{itemize}
    \item The paper does not release a new pretrained medical model, image generator, scraped dataset, or patient-level medical dataset.
    \item The submission does not include high-risk deployable assets.
    \item Restricted raw medical data are not redistributed.
\end{itemize}

\item \textbf{Licenses for existing assets}

\textbf{Question:} Are existing assets properly credited and are their licenses and terms respected?

\textbf{Answer:} \answerYes{}

\textbf{Justification:}
\begin{itemize}
    \item The paper cites the original datasets, benchmarks, baselines, and model backbones used in the experiments.
    \item Access to medical datasets, benchmarks, and model checkpoints follows the corresponding providers' licenses and terms of use.
    \item The paper does not redistribute restricted raw medical data or proprietary model assets.
\end{itemize}

\item \textbf{New assets}

\textbf{Question:} Are new assets introduced in the paper well documented?

\textbf{Answer:} \answerNA{}

\textbf{Justification:}
\begin{itemize}
    \item The paper introduces a method and experimental analysis, but the submission does not release a new dataset, benchmark or model checkpoint.
    \item All experiments are conducted using existing datasets, benchmarks, and model backbones cited in the paper.
    \item The code and reproducibility materials will be released after the review process.
\end{itemize}

\item \textbf{Crowdsourcing and research with human subjects}

\textbf{Question:} Does the paper include required information for crowdsourcing or human subject experiments?

\textbf{Answer:} \answerNA{}

\textbf{Justification:}
\begin{itemize}
    \item The paper does not conduct crowdsourcing experiments, recruit participants, or collect new annotations from human subjects.
    \item All experiments are conducted on existing datasets, benchmarks, and model outputs.
\end{itemize}

\item \textbf{Institutional review board approvals or equivalent}

\textbf{Question:} Does the paper describe IRB approvals or equivalent review for human subject research?

\textbf{Answer:} \answerNA{}

\textbf{Justification:}
\begin{itemize}
    \item The paper does not involve new patient data collection, participant recruitment, clinical intervention, or newly conducted human-subject research.
    \item The study uses existing datasets and benchmarks and follows their corresponding access terms.
    \item Therefore, IRB approval or equivalent review is not applicable.
\end{itemize}

\item \textbf{Declaration of LLM usage}

\textbf{Question:} Does the paper describe LLM usage if it is part of the core method?

\textbf{Answer:} \answerYes{}

\textbf{Justification:}
\begin{itemize}
    \item The paper fine-tunes and evaluates medical vision-language models, including Qwen3-VL-4B and InternVL3.5-4B.
    \item These models are used for gradient-based scoring, subset selection, fine-tuning, benchmark evaluation, and perturbation audits.
    \item The paper also uses Qwen3-4B for training-pool preprocessing, including data denoising, format checking, and task classification where applicable.
    \item All model-based preprocessing is applied only to training data. External benchmark labels, test answers, and model predictions are not used for subset construction.
\end{itemize}

\end{enumerate}

\end{document}